\documentclass[twoside,11pt]{article}
\usepackage{listings}
\usepackage[svgnames,dvipsnames]{xcolor} 
\usepackage{etoolbox}
\usepackage{jmlr2e}
\usepackage{url}
\usepackage{tikz}
\usepackage{pgfplots}
\usepackage{bm}
\usepackage{amsmath,amssymb}
\usepackage{natbib}
\usepackage[nolist]{acronym}
\usepackage[framemethod=TikZ]{mdframed}
\usepackage{graphicx}
\usepackage[abs]{overpic}
\usepackage{caption}
\usepackage{subcaption}

\usepackage{fancyvrb}
\usepackage{placeins}

\usepackage{mdframed}

\usepackage[inline]{asymptote}
\usepackage{mathtools}
\usepackage{inconsolata}
\usepackage{placeins}
\usepackage{setspace}

\usepackage{algorithm}
\usepackage{algpseudocode}

\usepackage[inline]{asymptote}
\usetikzlibrary{arrows}

\usepackage{enumitem}

\usepackage{multicol}

\begin{asydef}
defaultpen(fontsize(11));
\end{asydef}

\usetikzlibrary{tikzmark}
\usetikzlibrary{bayesnet}
\usetikzlibrary{pgfplots.groupplots}

\newcommand{\gpmem}{\texttt{gpmem}}
\newcommand{\emu}{{\textrm{emu}}}

\newcommand{\true}{{\textrm{true}}}
\newcommand{\rmnew}{{\textrm{new}}}

\newcommand{\noisy}{{\textrm{noisy}}}
\newcommand{\noise}{{\textrm{noise}}}
\newcommand{\sigmanoise}{\sigma_{\text{noise}}}
\newcommand{\Acal}{\mathcal{X}}
\newcommand{\R}{\mathbb{R}}
\newcommand{\xprime}{\mathbf{x}^\prime}
\newcommand{\yprime}{\mathbf{y}^\prime}
\newcommand{\yprimetop}{\mathbf{y}^{\prime \top}}
\newcommand{\xstar}{\mathbf{x}^*}
\newcommand{\ystar}{\mathbf{y}^*}
\newcommand{\abf}{\mathbf{x}}

\newcommand{\hbf}{\mathbf{h}}
\newcommand{\rbf}{\mathbf{y}}
\newcommand{\wbf}{\mathbf{w}}
\newcommand{\xbf}{\mathbf{x}}
\newcommand{\Mbf}{\mathbf{M}}
\newcommand{\Dbf}{\mathbf{D}}
\newcommand{\ybf}{\mathbf{y}}
\newcommand{\Kbf}{\mathbf{K}}
\newcommand{\Lbf}{\mathbf{L}}
\newcommand{\Sbf}{\mathbf{S}}
\newcommand{\Ktheta}{\mathbf{K}_{\bm{\theta}}}
\newcommand{\ktheta}{k_{\bm{\theta}}}
\newcommand{\Kpost}{\mathbf{K}_{\bm{\theta}}^\text{post}}
\newcommand{\mupost}{\bm{\mu}_{\bm{\theta}}^\text{post}}
\newcommand{\thetabf}{\bm{\theta}}
\newcommand{\Omegabf}{\bm{\Omega}}
\newcommand{\midtheta}{\mid \bm{\theta}}

\newcommand{\mubf}{\bm{\mu}}
\newcommand{\Krv}{\bm{\mathcal{K}}}
\newcommand{\Klin}{K^\text{linear}}
\newcommand{\Kper}{K^\text{periodic}}
\newcommand{\klin}{k^\text{linear}}
\newcommand{\kwn}{k^\text{wn}}
\newcommand{\kper}{k^\text{periodic}}
\newcommand{\kse}{k^\text{se}}
\newcommand{\Kse}{K^\text{se}}
\newcommand{\Ksrv}{\bm{\mathcal{K}}}
\newcommand{\Struct}{\text{Struct}}
\newcommand{\Simplify}{\text{Simplify}}
\newcommand{\Parse}{\text{Parse}}
\newcommand{\Cont}{\text{Contains}}
\newcommand{\WN}{\text{WN}}
\newcommand{\WNK}{\text{WN}(\Ksrv)}
\newcommand{\LINK}{\text{LIN}(\Ksrv)}
\newcommand{\PERK}{\text{PER}(\Ksrv)}
\newcommand{\SEK}{\text{SE}(\Ksrv)}

\newcommand{\pn}[1]{\left( #1 \right)}
\newcommand{\bkt}[1]{\left[ #1 \right]}
\newcommand{\br}[1]{\left\{ #1 \right\}}
\newcommand{\abs}[1]{\left\lvert #1 \right\rvert}
\newcommand{\Ebkt}[2][]{\mathbb{E}_{#1}\bkt{#2}}
\newcommand{\mvert}{\ \middle\vert\ }

\newcommand{\bmat}[1]{\begin{bmatrix} #1 \end{bmatrix}}

\newcommand{\Ncal}{\mathcal{N}}
\newcommand{\ftt}{\texttt{f}}

\newcommand{\xtt}{\texttt{x}}

\newcommand{\proposal}{{\textrm{proposal}}}
\newcommand{\update}{{\textrm{update}}}
\newcommand{\search}{{\textrm{search}}}
\newcommand{\accept}{{\textrm{accept}}}

\newcommand{\avg}{{\textrm{avg}}}

\newcommand{\ttheta}{\vartheta}
\newcommand{\muhat}{\widehat{\mu}}

\newcommand{\propstd}{\texttt{propstd}}

\newcommand{\myparagraph}[1]{\paragraph{#1}\mbox{}\\}

\begin{acronym}
\acro{GP} {Gaussian Processes}
\acro{MAP} {Maximum a posteriori}
\acro{MCMC} {Markov Chain Monte Carlo}
\acro{MDP} {Markov Decision Processes} 
\acro{MH} {Metropolis-Hastings}
\acro{SP} {Stochastic Process}
\acro{CCF} {Cosmic Calibration Framework}
\end{acronym}

  \definecolor{mygreen}{rgb}{0,0.6,0}
  \definecolor{mygray}{rgb}{0.5,0.5,0.5}
  \definecolor{mymauve}{rgb}{0.58,0,0.82}
  \definecolor{mygreen}{rgb}{0,0.4,0}
  \definecolor{mypurple}{rgb}{0.38,0,0.83}
  \definecolor{myorange}{rgb}{0.75,0.3,0}
  \definecolor{myblue}{RGB}{76,114,176}

 \lstdefinelanguage{Venture}{
    alsoletter={\&,=,!,?}
    keywordstyle=\color{black},
    morecomment=[l]{//},
    commentstyle=\color{mygreen},
    keywordstyle=[2]\color{DarkRed},
    keywords=[2]{assume,predict,infer,observe,define,assume_list,sample,call_back,for},
    keywordstyle=[3]\color{myorange},
    keywords=[3]{if,then,else,lambda,tag,do,proc,repeat,run,pass,true,false,quote,default,all,one,begin,let,lte,letrec,set!},
    keywordstyle=[4]\color{mypurple},
    keywords=[4]{flip,normal,flip,bernoulli,uniform_continuous,uniform_structure,uniform_discrete,gamma,mh,rejection,array,max,length,list,first,second,gridsearch_argmax,apply,add_funcs,mult_funcs,subset,lookup,size,mem\&em,contains,stats,map,mapv,gpmem,quadmem,make_whitenoise, mem,make_gp,make_const_func,make_squaredexp,draw_gp_curves,exclude,closest_point,contents,set_contents,contents_changed?,contents_changed,mark_recently_changed,peek,increment,tail,repeatO,confine,return,print,get_neal_blackbox,get_neal_data_xs,get_data_xs,size,get_bayesopt_blackbox},
    literate=%
    *{0}{{{\color{DarkBlue}0}}}1
    {1}{{{\color{DarkBlue}1}}}1
    {2}{{{\color{DarkBlue}2}}}1
    {3}{{{\color{DarkBlue}3}}}1
    {4}{{{\color{DarkBlue}4}}}1
    {5}{{{\color{DarkBlue}5}}}1
    {6}{{{\color{DarkBlue}6}}}1
    {7}{{{\color{DarkBlue}7}}}1
    {8}{{{\color{DarkBlue}8}}}1
    {9}{{{\color{DarkBlue}9}}}1,
  }
\newdimen\linenumbersep

\newcommand{\linenumber}[1]{%
  \linenumbersep 4pt%
  \advance\linenumbersep\mdflength{innerleftmargin}%
  \advance\linenumbersep\mdflength{innerlinewidth}%
  \advance\linenumbersep\mdflength{middlelinewidth}%
  \advance\linenumbersep\mdflength{outerlinewidth}%
  \advance\linenumbersep\mdflength{linewidth}%
  \makebox[0pt][r]{{\rmfamily\tiny#1}\hspace*{\linenumbersep}}}
\small

\let\OldStatex\Statex
\renewcommand{\Statex}[1][3]{%
  \setlength\@tempdima{\algorithmicindent}%
  \OldStatex\hskip\dimexpr#1\@tempdima\relax}

\ShortHeadings{Probabilistic Programming with Gaussian Processes}{}
\firstpageno{1}

\begin{document}
\title{Probabilistic Programming with Gaussian Process Memoization}

\author{\name Ulrich Schaechtle \email ulrich.schaechtle@rhul.ac.uk \\
	      \addr Department of Computer Science\\
              Royal Holloway, University of London
       \AND \name Ben Zinberg \email bzinberg@alum.mit.edu \\
              \addr Computer Science and Artificial Intelligence Laboratory\\
              Massachusetts Institute of Technology
       \AND \name Alexey Radul \email axch@mit.edu \\
              \addr Computer Science and Artificial Intelligence Laboratory\\
              Massachusetts Institute of Technology
       \AND \name Kostas Stathis \email kostas.stathis@rhul.ac.uk\\
              \addr Department of Computer Science\\
       Royal Holloway, University of London
       \AND \name Vikash K. Mansinghka \email vkm@mit.edu \\
	      \addr Computer Science and Artificial Intelligence Laboratory\\
	      Massachusetts Institute of Technology
} 

       \editor{N.A.}

\maketitle
\noindent\begin{abstract}
Gaussian Processes (GPs) are widely used tools in statistics, machine learning, robotics, computer vision, and scientific computation. However, despite their popularity, they can be difficult to apply; all but the simplest classification or regression applications require specification and inference over complex covariance functions that do not admit simple analytical posteriors. This paper shows how to embed Gaussian processes in any higher-order probabilistic programming language, using an idiom based on memoization, and demonstrates its utility by implementing and extending classic and state-of-the-art GP applications. The interface to Gaussian processes, called \gpmem, takes an arbitrary real-valued computational process as input and returns a statistical emulator that automatically improve as the original process is invoked and its input-output behavior is recorded.  The flexibility of \gpmem\ is illustrated via three applications: (i) Robust GP regression with hierarchical hyper-parameter learning, (ii) discovering symbolic expressions from time-series data by fully Bayesian structure learning over kernels generated by a stochastic grammar, and (iii) a bandit formulation of Bayesian optimization with automatic inference and action selection. All applications share a single 50-line Python library and require fewer than 20 lines of probabilistic code each.
\end{abstract}


\section{Introduction}
\acl{GP} (GPs)  are widely used tools in statistics~\citep{barry1986nonparametric}, machine learning~\citep{neal1995bayesian,williams1998bayesian,kuss2005assessing,rasmussen2006gaussian,damianou2013deep}, robotics \citep{ferris2006gaussian}, computer vision~\citep{kemmler2013one}, and scientific computation~\citep{kennedy2001bayesian,schneider2008simulations,kwan2013cosmic}.
They are also central to probabilistic numerics, an emerging effort to develop more computationally efficient numerical procedures, and to Bayesian optimization, a family of meta-optimization techniques that are widely used to tune parameters for deep learning algorithms~\citep{snoek2012practical,gelbart2014bayesian}. 
They have even seen use in artificial intelligence. For example, by searching
over structured kernels generated by a stochastic grammar, the ``Automated
Statistician" system can produce symbolic descriptions of time series
~\citep{duvenaud2013structure} that can be translated into natural
language~\citep{lloyd2014automatic}.

This paper shows how to integrate \acsp{GP} into higher-order
probabilistic programming languages and illustrates the utility of this
integration by implementing it for the Venture platform. The key idea is to use
GPs to implement a kind of “statistical” or “generalizing” memoization. The
resulting higher-order procedure, called {\tt gpmem}, takes a kernel function
and a source function and returns a GP-based statistical emulator for the source
function that can be queried at locations where the source function has not yet
been evaluated. When the source function is invoked, new datapoints are
incorporated into the emulator. In principle, the covariance function for the
GP is also allowed to be an arbitrary probabilistic program. This simple packaging covers the full range of uses of the GP described above, including both statistical applications and applications to scientific computation and uncertainty quantification.

This paper illustrates {\tt gpmem} by embedding it in Venture, a general-purpose, higher-order probabilistic programming platform~\citep{mansinghka2014venture}. Venture has several distinctive capabilities that are needed for the applications in this paper. First, it supports a flexible foreign interface for modeling components that supports the efficient rank-1 updates required by standard GP implementations. Second, it provides inference programming constructs that can be used to describe custom inference strategies that combine elements of gradient-based, Monte Carlo, and variational inference techniques. This level of control over inference is key to state-of-the-art applications of GPs. Third, it supports models with stochastic recursion, a priori unbounded support sets, and higher-order procedures; together, these enable the combination of stochastic grammars with a fast GP implementation, needed for structure learning. Fourth, Venture permits nesting of modeling and inference, which is needed for the use of GPs in Bayesian optimization over general objective functions that may in general themselves be derived from modeling and inference.

To the best of our knowledge, this is the first general-purpose integration of
\acsp{GP} into a probabilistic programming language. Unlike software
libraries such as GPy~\citep{gpy2014}, our embedding allows uses of GPs that go beyond classification and regression to include state-of-the art applications in structure learning and meta-optimization.

This paper presents three applications of gpmem: (i) a replication of results by~\citet{neal1997monte} on outlier rejection via hyper-parameter inference; (ii) a fully Bayesian extension to the Automated Statistician project; and (iii) an implementation of Bayesian optimization via Thompson sampling. The first application can in principle be replicated in several other probabilistic languages embedding the proposal that is described in this paper. The remaining two applications rely on distinctive capabilities of Venture: support for fully Bayesian structure learning and language constructs for inference programming. All applications share a single 50-line Python library and require fewer than 20 lines of probabilistic code each.


\setcounter{figure}{0}
\section{Background on Gaussian Processes}
\acused{GP}
Gaussian Processes (GPs) are a Bayesian method for regression. We consider the regression input
to be real-valued scalars $x_i$ and the regression output $f(x_i)=y_i$ as the value of a
function $f$ at $x_i$. The complete training data will be denoted by column
vectors $\mathbf{x}$ and $\mathbf{y}$. Unseen test input is denoted with
$\xprime$.
\ac{GP}s present a non-parametric way to express prior knowledge on the space of all possible functions $f$ modeling
a regression relationship.
Formally, a GP is an infinite-dimensional extension of the multivariate Gaussian distribution.

The collection of random variables $\br{f(x_i)=y_i}$ (indexed by $i$) represents the
values of the function $f$ at each location $x_i$.
We write $f \sim \mathcal{GP}(\mu,k \mid \thetabf_\text{mean},\thetabf)$, where
$\mu$ is the {\em mean function} and $k$ is the {\em covariance function} or {\em kernel}.
That is, $\mu(x_i \mid \thetabf_\text{mean})$ is the prior mean of the random variable $y_i$, and
$k(x_i,x_j \midtheta)$ is the prior covariance of the random variables $y_i$
and $y_j$. The output of both mean and covariance function are conditioned on a
few free hyper-parameters parameterizing $k$ and $\mu$. We refer to these
hyper-parameters as $\thetabf$ and
$\thetabf_\text{mean}$
respectively.
To simplify the calculation below, we will assume the prior mean $\mu$ is identically zero; once the derivation is done, this assumption can be easily relaxed via translation.
We use upper case
italic $K(\xbf,\xprime \midtheta )$ for a function that returns a matrix of dimension $I \times J$
with entries $k(x_i,x_j \midtheta )$ and with $x_i \in \xbf$ and $x_j \in
\xprime$ where $I$ and $J$ indicate the length of the column vectors $\xbf$ and
$\xprime$.
Throughout, we write $\Kbf_{(\thetabf,\xbf,\xprime)}$ for the prior covariance
matrix that results from computing  $K(\xbf,\xprime \midtheta )$. In the
following, we will sometimes drop the subscript $\xbf,\xprime$, writing
only $\Ktheta$, for clarity. Note that we do this only in cases when both
input vectors are identical and correspond to training input $\xbf$.
We differentiate two different situations leading to different ways samples can be generated with
this setup:
\begin{enumerate}
\item ${\yprime}$ - the predictive posterior sample from a distribution
conditioned on observed input $\xbf$ with observed output $\ybf$ and conditioned on
$\thetabf$.
\item $\ystar$ - a sample from the predictive prior. We will describe
situations, where the \ac{GP} has not seen any data $\xbf,\ybf$ yet. In this
case, we sample from a Gaussian distribution with 
     $\ystar \sim \mathcal{N}\big(0,K(\xstar, \xstar \midtheta
)\big)$; where the symbol $*$ indicates that no data has been observed yet. 
\end{enumerate}

We now show how to compute the predictive posterior distribution of test output $\yprime := f(\xprime)$ conditioned on training data $\ybf := f(\xbf)$.  (Here $\xbf$ and $\xprime$ are known constant vectors, and we are conditioning on an observed value of $\ybf$.) The predictive posterior can be computed by first forming the joint density when both training and test data are treated as randomly chosen from the prior, then fixing the value of $\ybf$ to a constant.  To start, let
\[
  \Sigma := \bmat{
    K(\xbf, \xbf \midtheta )     & K(\xbf, \xprime \midtheta )     \\
    K(\xprime, \xbf \midtheta ) & K(\xprime, \xprime \midtheta )
  }
  \text{ and }
  \Sigma^{-1} =: \bmat{
    \Mbf_{11} & \Mbf_{12} \\
    \Mbf_{21} & \Mbf_{22}
  }.
\]
We then have
\[
  P(\ybf, \yprime \midtheta )
  \propto
  \exp\br{
    -\frac12
    \bmat{\ybf^\top & \yprimetop}
    \bmat{\Mbf_{11} & \Mbf_{12} \\ \Mbf_{21} & \Mbf_{22}}
    \bmat{\ybf \\ \yprime}
  }.
\]
Treating $\ybf$ as a fixed constant, we obtain
\[
  P\pn{\yprime \mvert \ybf,\thetabf}
  \propto
  P(\ybf, \yprime \midtheta )
  \propto
  \exp\br{
    -\frac12 \yprimetop \Mbf_{22} \yprime
    - \hbf^\top \yprime
  },
\]
where $\hbf = M_{21} \ybf$ is a constant vector.  Thus $P(\yprime | \ybf,
\thetabf)$ is Gaussian,
\begin{equation}\label{eq:pred_posterior}
  P\pn{\yprime \mvert \ybf, \thetabf} \sim \Ncal(\mupost,
\Kbf_{\thetabf}^\text{post}),
\end{equation}
with covariance matrix $\Kpost = \Mbf_{22}^{-1}$.  To find its mean $\mupost$,
we note that $P_{\yprime|\ybf,\thetabf}(\yprime + \mupost)$ is Gaussian with
the same covariance as $P(\yprime | \ybf,\thetabf)$, but its exponent has no linear term:
\begin{align*}
  P_{\yprime|\ybf,\thetabf} \pn{\yprime + \mupost \mvert \ybf,\thetabf}
  &\propto
  \exp\br{
    -\frac12 (\yprime + \mupost)^\top \Mbf_{22} (\yprime + \mupost)
    - \hbf^\top (\yprime + \mupost)
  } \\
  &\propto
  \exp\br{
    -\frac12 \yprime\top \Mbf_{22} \yprime
    - \underbrace{(\hbf + \Mbf_{22} \mupost)^\top}_{\text{must be $0$}} \yprime
  }.
\end{align*}
Thus $\hbf = -\Mbf_{22} \mupost$ and $\mupost = -\Mbf_{22}^{-1} \hbf =
-\Mbf_{22}^{-1} \Mbf_{21} \ybf$.

The partioned inverse equations (\citealp*{barnett1979matrix} following \citealp*{mackay1998introduction}) give
\begin{align*}
  \Mbf_{22} &= \big(K(\xprime,\xprime \midtheta ) - K(\xprime,\xbf \midtheta ) K(\xbf,\xbf \midtheta )^{-1}
K(\xbf,\xprime \midtheta )\big)^{-1}, \\
  \Mbf_{21} &= -\Mbf_{22} K(\xprime,\xbf \midtheta ) K(\xbf,\xbf \midtheta )^{-1}.
\end{align*}
Substituting these in the above gives
\begin{align}
  \Kpost &= K(\xprime,\xprime \midtheta ) - K(\xprime,\xbf \midtheta )
K(\xbf,\xbf \midtheta )^{-1} K(\xbf,\xprime \midtheta ),\label{eq:K_hat} \\
  \mupost &= K(\xprime,\xbf \midtheta ) K(\xbf,\xbf \midtheta )^{-1}\ybf.\label{eq:mu_hat}
\end{align}
Together, $\mupost$ and $\Kpost$ determine the computation of the predictive posterior
with unseen input data (\ref{eq:pred_posterior}).

Often one assumes the observed regression output is noisily measured, that is,
one only sees the values of $\ybf_\noisy = \ybf + \wbf$ where $\wbf$ is
Gaussian white noise with variance $\sigma_\noise^2$. This noise term can be
absorbed into the covariance function $K(\xbf,\xbf \midtheta)$.
The log-likelihood of a \ac{GP} can then be written as:
\begin{equation}
\label{eq:gplogdens}
\log P(\ybf \mid \xbf,\thetabf) =
-\frac12 \ybf^\top 
\Ktheta^{-1} \ybf
- \frac12\log \abs{\Ktheta}
- \frac{n}{2}\log 2\pi
\end{equation}
where $n$ is the number of data points.
Both log-likelihood and predictive posterior can be computed efficiently using a \ac{SP} in Venture~\citep{mansinghka2014venture}
with an algorithm that resorts to Cholesky factorization\citep[chap. 2]{rasmussen2006gaussian}. 
We write the Cholesky factorization as 
$\Lbf\coloneqq \text{chol}(\Ktheta)$ when
:
\begin{equation}
\Ktheta = \Lbf\Lbf^\top
\end{equation}
where $\Lbf$ is a lower triangular matrix. This allows us to compute the inverse of a covariance matrix as
\begin{equation}
\Ktheta^{-1} = (\mathbf{L}^{-1})^\top (\mathbf{L}^{-1})
\end{equation}
and its determinant as 
\begin{equation}
det(\Ktheta) = det(\mathbf{L})^2
\end{equation}
We compute (\ref{eq:gplogdens}) as
\begin{equation}
\log(P(\ybf \mid \xbf, \thetabf)\coloneqq - \frac{1}{2} \ybf^\top \bm{\alpha} - \sum_i \log \mathbf{L}_{ii} - \frac{n}{2} \log 2 \pi
\end{equation}
where 
\begin{equation}
\label{eq:chol_L}
\mathbf{L} \coloneqq \text{chol}(\Ktheta)
\end{equation}
and 
\begin{equation}
\label{eq:alpha}
\bm{\alpha} \coloneqq  \mathbf{L}^\top \backslash(\mathbf{L} \backslash \ybf). 
\end{equation}
This results in a computational complexity for sampling in the number of data points of $O(n^3/6)$ for (\ref{eq:chol_L}) an $O(n^2/2)$ for (\ref{eq:alpha}). 
Above, we defined the \ac{GP} prior as $\ystar \sim
\mathcal{N}\big(0,K(\xbf_*,\xbf_* \midtheta )\big)$.
We see that this prior is fully determined by its covariance function.
\subsection{Covariance Functions}
The covariance function (or kernel) of a \ac{GP} governs high-level properties
of the observed data such as smoothness or linearity. The high-level properties
are indicated with superscript on functions.  A linear covariance can be written as:
\begin{equation}\label{eq:LIN1}
    \klin =   \sigma_1^2(x x^\prime).
\end{equation}
We can also express periodicity:
\begin{equation}\label{eq:PER1}
    \kper =  \sigma_2^2 \exp \bigg( \frac{2 \sin^2 ( \pi (x - x^\prime)/p}{\ell^2} \bigg). 
\end{equation}
By changing these properties we get completely different prior behavior for sampling $\ystar$ from a
\ac{GP} with a linear kernel
\[
\ystar \sim \mathcal{N}\big(0,\Klin(\xstar,\xstar \mid \sigma_1 )\big)
\]
as compared to sampling from the prior predictive with a periodic kernel (as depicted in 
Fig. \ref{fig:composition_tutorial} (c) and (d))
\[
\ystar \sim \mathcal{N}\big(0,\Kper(\xstar,\xstar \mid \sigma_2, \ell )\big).
\]
\begin{figure}
\input{figs/composition/composition.tex}
\setcounter{figure}{0}
\caption{\footnotesize We depict kernel composition. 
(a) shows raw data (black) generated with a sine function with linearly growing amplitude (blue).
This data is used for all the plots (c-h). 
(b) shows the linear and the periodic base kernel in functional form as well as a composition of both. 
The multiplication of the two kernels indicates local interaction. The local interaction we account for in this case is the growing amplitude (a). For each column (c-h) $\bm{\theta}$ is different.(c-e) show samples from the prior
predictive $\ystar$ where random parameters are used, that is, we sample before any data points are observed.
(f-h) show samples from the predictive posterior $\yprime$, after the data has been observed.}
\label{fig:composition_tutorial}
\end{figure}
These high-level properties are compositional via addition and multiplication of different covariance functions. 
That means that we can also combine these properties.
By using multiplication of kernels we can model a local interaction of two components, for example 
\begin{equation}\label{eq:LINxPER}
    \klin \times \kper =  \sigma_1^2(x x^\prime)\, \sigma_2^2 \exp \bigg(
\frac{2 \sin^2 ( \pi (x - x^\prime)/p}{\ell^2} \bigg).
\end{equation}
This results in a combination of the higher level properties of linearity and  periodicity.
In Fig \ref{fig:composition_tutorial} (e) we depict samples for $\ystar$ that are periodic
with linearly increasing amplitude.
We consider this a local interaction because the actual interaction depends on the similarity
of two data points.
An addition of covariance functions models a global interaction, that is an interaction of two high-level components that is qualitatively not dependent on the input space. An example for this a periodic function with a linear
trend.

For each kernel type, each $\bm{\theta}$ is different, that is, in (\ref{eq:LIN1}) we have $\thetabf=\{\sigma_1\}$,
in (\ref{eq:PER1}) we have $\bm{\theta}=\{\sigma_2,p,\ell\}$ and in 
(\ref{eq:LINxPER}) we have $\bm{\theta}=\{\sigma_1,\sigma_2,p,\ell\}$.
Adjusting these hyper-parameters changes lower level qualitative attributes such as length
scales ($\ell$) while preserving the higher level qualitative properties of the distribution
such as linearity.

If we choose a suitable set of hyper-parameters, for example by performing inference, we
can capture the underlying dynamics of the data well (see Fig.
\ref{fig:composition_tutorial} (f-h)) while sampling $\yprime$.
Note that goodness of fit is not only limited to the parameters. A too simple qualitative structure
implies unsuitable behaviour, as for example in (Fig. \ref{fig:composition_tutorial} (g)) where additional 
recurring spikes are introduced to account for the changing amplitude of the true function that 
generated the data.


\section{Gaussian Process Memoization in Venture}
Memoization is the practice of storing previously computed values of a function so that future calls with the same inputs can be evaluated by lookup rather than re-computation.
To transfer this idea to probabilistic programming, we now introduce a language construct called a
\emph{statistical memoizer}.  Suppose we have a function $\ftt$ which can be evaluated 
but we wish to learn about the behavior of $\ftt$ using as
few evaluations as possible.  The statistical memoizer, which here we give the
name \gpmem, was motivated by this purpose.  It produces two outputs:
\[ \ftt \xrightarrow{\gpmem} (\ftt_{\text{compute}}, \ftt_\emu). \]
The function $\ftt_{compute}$ calls $\ftt$ and stores the output in a memo
table, just as traditional memoization does.  The function $\ftt_\emu$ is
an online statistical emulator which uses the memo table as its training
data.  A fully Bayesian emulator, modelling the true function $\ftt$ as a
random function $f \sim P(f)$, would satisfy
\[
\texttt{(}\ftt_\emu\ \xtt_1\ \ldots\ \xtt_k\texttt{)}
\sim
P\pn{
  f(\xtt_1), \ldots, f(\xtt_k)
  \mvert
  \text{$f(\xtt) = \texttt{(f x)}$ for each $\xtt$ in memo table}
}.
\]
Different implementations of the statistical memoizer can have
different prior distributions $P(f)$; in this paper, we deploy a \ac{GP} 
prior (implemented as \texttt{gpmem} below).  Note that we require the ability
to sample $\ftt_\emu$ jointly at multiple inputs because the values of
$f(\xtt_1),\ldots,f(\xtt_k)$ will in general be dependent.

\begin{figure}
\noindent\input{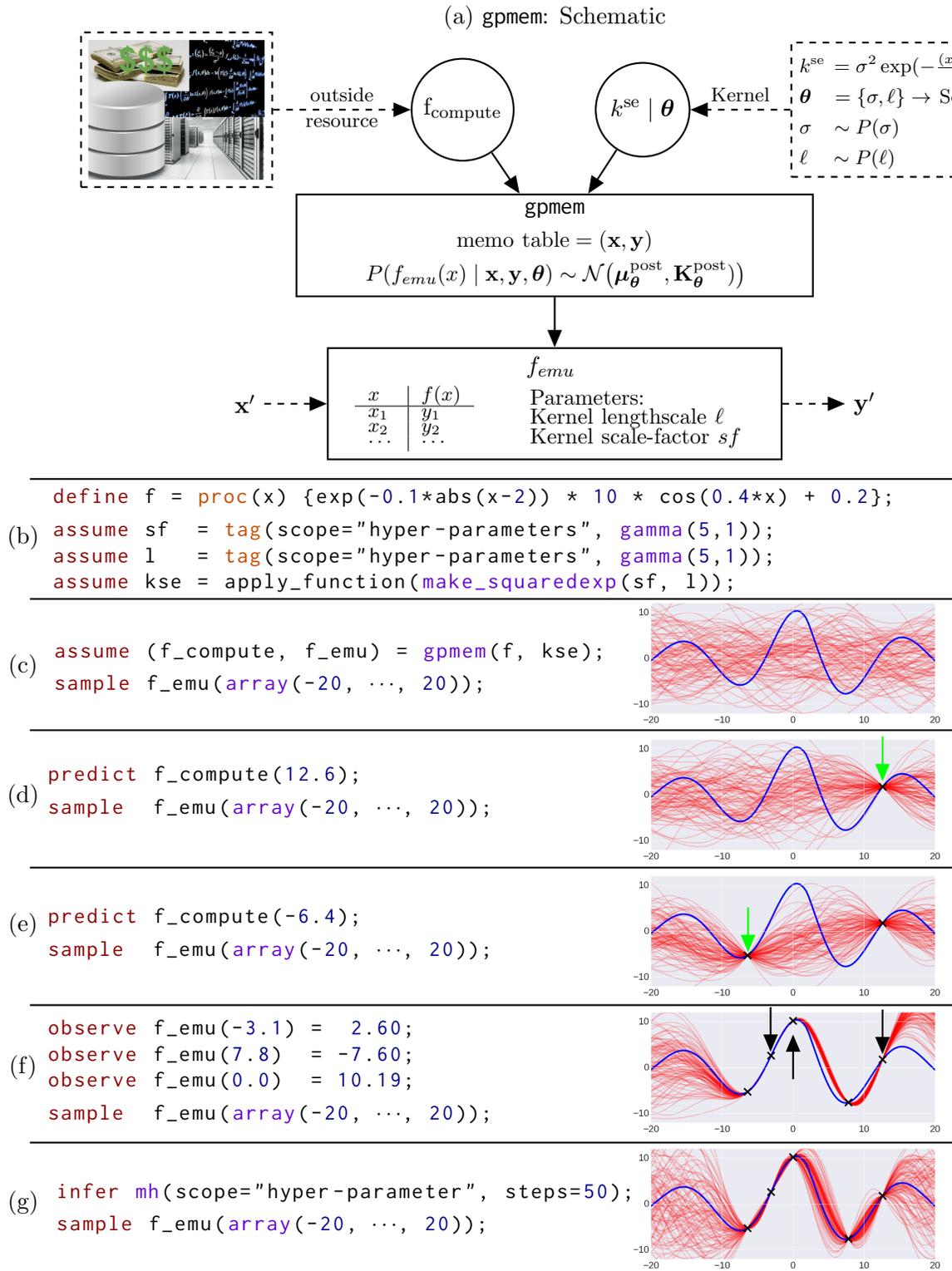}
\captionsetup{aboveskip=-3pt}
\caption{\footnotesize \gpmem\ tutorial. The top shows a schematic of \gpmem.
  \texttt{f\_compute} probes an outside resource.
  This can be expensive (top left).
  Every probe is memoized and improves the emulator. Below the schematic we see the evolution
  of \gpmem's state of believe of the world given certain Venture
  directives. On the right, we depict the true function (blue), samples from the
emulator (red) and incorporated observations (black).}
\label{fig:gpmem_tutorial}
\end{figure}

We explain how \gpmem, the statistical memoizer with \ac{GP}-prior, works using a simple tutorial
(Fig. \ref{fig:gpmem_tutorial}). 
The top panel (Fig. \ref{fig:gpmem_tutorial}, (a)) of this figure sketches the schematic of \gpmem.
$\ftt$ is the external process that we memoize. It can be evaluated using resources that potentially come
from outside of Venture.  
We feed this function into \gpmem\ alongside
a parameterised kernel $k$.  
In this example, we make the qualitative assumption of $f$ being smooth, and define
$k$ to be a squared-exponential covariance function:
\[
k = \kse = \sigma^2 \exp(-\frac{(x-x^\prime)^2}{2\ell^2}).
\]
The hyper-parameters $\thetabf$ for this kernel are sampled from a 
prior distribution which is depicted in the top right box.
Note that we annotate $\bm{\theta}=\{\texttt{sf},\texttt{l}\}$ for subsequent
inference as belonging to the scope ``hyper-parameter".

\gpmem\ implements a memoization table, where all previously
computed function evaluations ($\{\xbf,\ybf\}$) are stored. We also initialize a \ac{GP}-prior that
will serve as our statistical emulator:
\[
P(f_{emu}(x) \mid \xbf,\ybf,\thetabf)\sim
\mathcal{N}\big(\mupost,\Kpost)\big)
\]
where 
\[
P(f_{emu}(x) \mid \mathbf{x},\ybf,\thetabf) = \yprime 
\]
under the traditional \ac{GP} perspective.
All value pairs stored in the memoization table ($\text{memo table} = (\xbf,\ybf)$) are incorporated as observations of
the \ac{GP}.
We simply feed the regression input
into the emulator and output a predictive posterior Gaussian distribution determined by the \ac{GP} and
the memoization table.

We can either define the function f that serves as as input for \gpmem\
 natively in Venture
(as shown in the Fig. \ref{fig:gpmem_tutorial} (b)) or we interleave Venture with foreign code. 
This can be useful when $\ftt$ is computed with the help of outside resources.
We define and parameterize a squared-exponential kernel (b) which we then supply to
\gpmem\ (Fig. \ref{fig:gpmem_tutorial} (c)).
Before making any observations or calls to $\ftt$
we can sample from the prior at the inputs from -20 to 20 using the emulator :
    \begin{lstlisting}
    assume (f_compute, f_emu) = gpmem(f, kse));

    sample f_emu(array(-20, ..., 20));
    \end{lstlisting}
where the second line corresponds to:
\[ 
\ystar \sim \mathcal{N}\Bigg(0,\Kse\bigg(
\bmat{
-20 \\
\cdots \\
20
},
\bmat{
-20 \\
\cdots \\
20
}
\mid \thetabf=\{\sigma,\ell\}
\bigg)
\Bigg).
\]
In Fig. \ref{fig:gpmem_tutorial} (d), we probe the external function $\ftt$ at point 12.6 and memoize its result by calling 
   \begin{lstlisting}
    predict f_compute (12.6);
    \end{lstlisting}
When we subsequently sample from the emulator, that is compute the $\yprime$ at the input
$\xprime= \bmat{-20, \cdots, 20}^\top$, we see how the posterior shifts from uncertainty to near certainty close to the input 12.6.

We can repeat the process at a different point (probing point -6.4 in Fig.
\ref{fig:gpmem_tutorial} (e)) to see that we gain certainty about another part of the curve. 

We can add information to $\texttt{f}_\text{emu}$ about presumable value pairs of $\ftt$ without calling $\texttt{f}_\text{compute}$
(Fig. \ref{fig:gpmem_tutorial} (f)).
If a friend tells us the value of $\ftt$ we can call $\texttt{observe}$ to store this information in the incorporated observations for $\texttt{f}_\text{emu}$ only:
    \begin{lstlisting}
    observe f_emu( -3.1) = 2.60;
    \end{lstlisting}
We have this value pair now available for the computation $\yprime$. 
For sampling with the emulator, the effect is the same as calling predict with the $\texttt{f}_\text{compute}$.
However, we can imagine at least one scenario where such as distinction in the treatment of observations 
is beneficial. Let us say we do not only have the real function available but also a domain expert with knowledge 
about this function.
This expert could tell us what the value is at a given input.
Potentially, the value provided by the expert could disagree with the value computed with $\ftt$ for example 
due to different levels of observation noise. 

Finally, we can update our posterior by inferring the posterior over hyper-parameter values $\thetabf$.
For this we use the defined scopes, which tag a collection of related random choices, such
as all hyper-parameters $\thetabf$.
These tags are supplied to the
inference program (in this case, MH) to specify on which random variables inference
should be done:
    \begin{lstlisting}
    infer mh(scope="hyper-parameters", steps=50);
    \end{lstlisting}
In this case, we perform one \ac{MH} transition over the scope hyper-parameters
and choose a random member of this scope, that is we choose one hyper-parameter at random.
We can also define custom inference actions. Let's define \ac{MH} with Gaussian
drift proposals.
    \begin{lstlisting}
    define drift_kernel = proc(x) { normal(x, 1) };

    define my_markov_chain =
	apply_mh_correction(
	    subproblem=choose_by_scope("hyper-parameters"),
	    proposal=symmetric_local_proposal_from_chain(drift_kernel))

    infer my_markov_chain;
    \end{lstlisting}
Note that this inference is not in the Figure. The important part of the above code snippet is \texttt{drift\_kernel}, which is where we say 
that at each step of our Markov chain, we would like to propose a transition by sampling
a new state from a unit normal distribution whose mean is the current state. 

The newly inferred hyper-parameters allow us now to adequately reflect uncertainty
about the curve given all incorporated observations (compare
Fig. \ref{fig:gpmem_tutorial}, bottom panel (g) on  the right with the samples
before inference, one panel above (f)).

\section{Applications}
This paper illustrates the flexibility of \gpmem\ by showing how it can concisely encode three different applications of \ac{GP}s.
The first is a standard example from hierarchical Bayesian statistics, where Bayesian inference over a hierarchical hyper-prior is used to provide a curve-fitting methodology that is robust to outliers.
The second is a structure learning application from probabilistic artificial intelligence, where \ac{GP}s are used to discover qualitative structure in time series data.
The third is a reinforcement learning application, where \ac{GP}s are used as part of a Thompson sampling formulation of Bayesian optimization for general real-valued objective functions with real inputs.

\subsection{Nonlinear regression in the presence of outliers}
We can apply \gpmem\ for regression in a hierarchical Bayesian setting
(Fig. \ref{fig:neal_tutorial}).  
\begin{figure}
\input{figs/neal_tutorial.tex}
\captionsetup{aboveskip=-7pt}
\caption{\footnotesize Regression with outliers and hierarchical prior
structure.}
\label{fig:neal_tutorial}
\end{figure}
In a Bayesian treatment of  hyper-parameter learning for \ac{GP}s,
we can write the posterior probability of the hyper-parameters of a GP  (Fig.
\ref{fig:neal_tutorial}, (a)) given covariance function $\Krv=k$ as:
\begin{equation}
\label{eq:hyperProbability}
P(\thetabf=\{sf,\ell,\sigma\} \mid \mathbf{D},\Krv ) = \frac{P(\mathbf{D} \mid
\thetabf, \Krv)P(\thetabf \mid \Krv) }{P(\mathbf{D} \mid
\Krv)}
\end{equation}
where $\mathbf{D} = \{\xbf, \ybf\}$ is a training data set and $\Krv$ is treated
as a random variable over covariance functions. Since we can apply
\gpmem\ to any process or procedure, it can be used in situations where a data
set is available only via a look-up function $\texttt{f\_look\_up}$.
In fact, we demonstrate \gpmem's application to regression using an example where
the data was generated by a function which is not available, that is, we do not
provide the synthetic function to \gpmem\ but only a data set (Fig.
\ref{fig:neal_tutorial} (b)).
This function, $f_\text{true}$, is taken from a paper on the
treatment of outliers with hierarchical Bayesian hyper-priors for
\ac{GP}s~\citep{neal1997monte}:
\begin{equation}
f_\text{true}(x) =  0.3 + 0.4 x + 0.5 \sin(2.7x) + \frac{1.1}{(1+ x^2)} + \eta
\;\;\; with\;\;\eta \sim \mathcal{N}(0,\sigmanoise).
\end{equation}
We synthetically generate outliers by setting $\sigmanoise = 0.1$ in $95\%$ of
the cases and to $\sigmanoise = 1$ in the remaining cases. 
Instead of accessing the $f_\text{true}$ directly, we are accessing the $\texttt{data}$ in form of
a a two dimensional $\texttt{array}$ with $\texttt{f\_look\_up}$.

We set $\Krv = k^{\text{se}+\text{wn}}$ and parameterize it with $\bm{\theta}=\{sf,\ell,\sigma\}$.
For these hyper-parameters, Neals work suggests a hierarchical system for
hyper-parameterization.
Here, we draw hyper-parameters from $\Gamma$ distributions:
\begin{equation}
\label{eq:hyper-ell}
\ell \sim \Gamma(\alpha_1,\beta_1),\;\sigma \sim \Gamma(\alpha_2,\beta_2)
\end{equation} 
and in turn sample the $\alpha$ and $\beta$ from $\Gamma$ distributions as well:
\begin{equation}
\label{eq:hyper-alpha}
\alpha_1 \sim \Gamma(\alpha^1_{\alpha},\beta^1_{ \alpha } ),\; \alpha_2 \sim \Gamma(\alpha^2_{\alpha},\beta^2_{\alpha}),\cdots
\end{equation}
We model this in Venture as illustrated in Fig. \ref{fig:neal_tutorial} (c),
using the build-in \ac{SP} $\texttt{gamma}$. 

In Fig. \ref{fig:neal_tutorial}, panel (d), we see that $k^{\text{se}+\text{wn}}$
is defined as a composite covariance function. It is the sum ($\texttt{add\_funcs}$) of
a squared exponential kernel ($\texttt{make\_squaredexp}$) and a white noise
($\kwn$, Appendix A)
kernel which is implemented with $\texttt{make\_whitenoise}$\footnote{Note
that in Neal's work \citeyearpar{neal1997monte} the sum of an SE
plus a constant kernel is used. We use a WN kernel for illustrative purposes
instead.}. 
We then initialize \gpmem\ feeding it with $\texttt{composite\_covariance}$ and the data
look-up function $\texttt{f\_look\_up}$. 
We sample from the prior $\ystar$ with random parameters $\texttt{sf,l}$ and $\texttt{sigma}$ and 
without any observations available (Fig. \ref{fig:neal_tutorial}, panel (e)).
We depict those samples on the right (red), alongside the true function that generated the data (blue) and
the data points we have available in the data set (black).

We can incorporate observations using both \texttt{observe} and \texttt{predict} (Fig. \ref{fig:neal_tutorial} (f)).
When we subsequently sample $\yprime$ from the emulator with
$\mathcal{N}(\mupost,\Kpost)$, we can see that the \ac{GP} posterior incorporates knowledge 
about the $\texttt{data}$. Yet, the hyper-parameters $\texttt{sf,l}$ and $\texttt{sigma}$ are still
random, so the emulator does not capture the true underlying dynamics
($f_\text{true}$) of the \texttt{data} correctly. 

Next, we demonstrate how we can capture these underlying dynamics within only
100  nested \ac{MH} steps on the hyper-parameters to get a good approximation
for their posterior $\yprime$ (Fig. \ref{fig:neal_tutorial} (g)).
We say nested because we first take two sweeps in the scope
$\texttt{hyperhyper}$ which characterizes (\ref{eq:hyper-alpha}) and then one
sweep on the scope $\texttt{hyper}$ which characterizes (\ref{eq:hyper-ell}).
This is repeated 100 times using $\texttt{repeat( 100, do(}\cdots\;$.
Note that Neal devises an additional noise model and performs a large number of Hybrid-Monte Carlo and Gibbs steps to achieve this, whereas inference in Venture with \gpmem\ is merely one line of code. 

Finally, we can change our inference strategy altogether. If we decide that instead of
following a Bayesian sampling approach, we would like to perform empirical optimization,
we do this by only changing one line of code, deploying $\texttt{gradient-ascent}$ instead
of $\texttt{mh}$ (Fig. \ref{fig:neal_tutorial} (h)).

\subsection{Discovering qualitative structure from time series data}\label{sec:structurelearning}
Inductive learning of symbolic expressions for continuous-valued time series
data is a hard task which has recently been tackled using a greedy search over 
the approximate posterior of the possible kernel compositions for
\ac{GP}s~\citep{duvenaud2013structure,lloyd2014automatic}\footnote{\url{http://www.automaticstatistician.com/}}.

\begin{figure}
\centering
\input{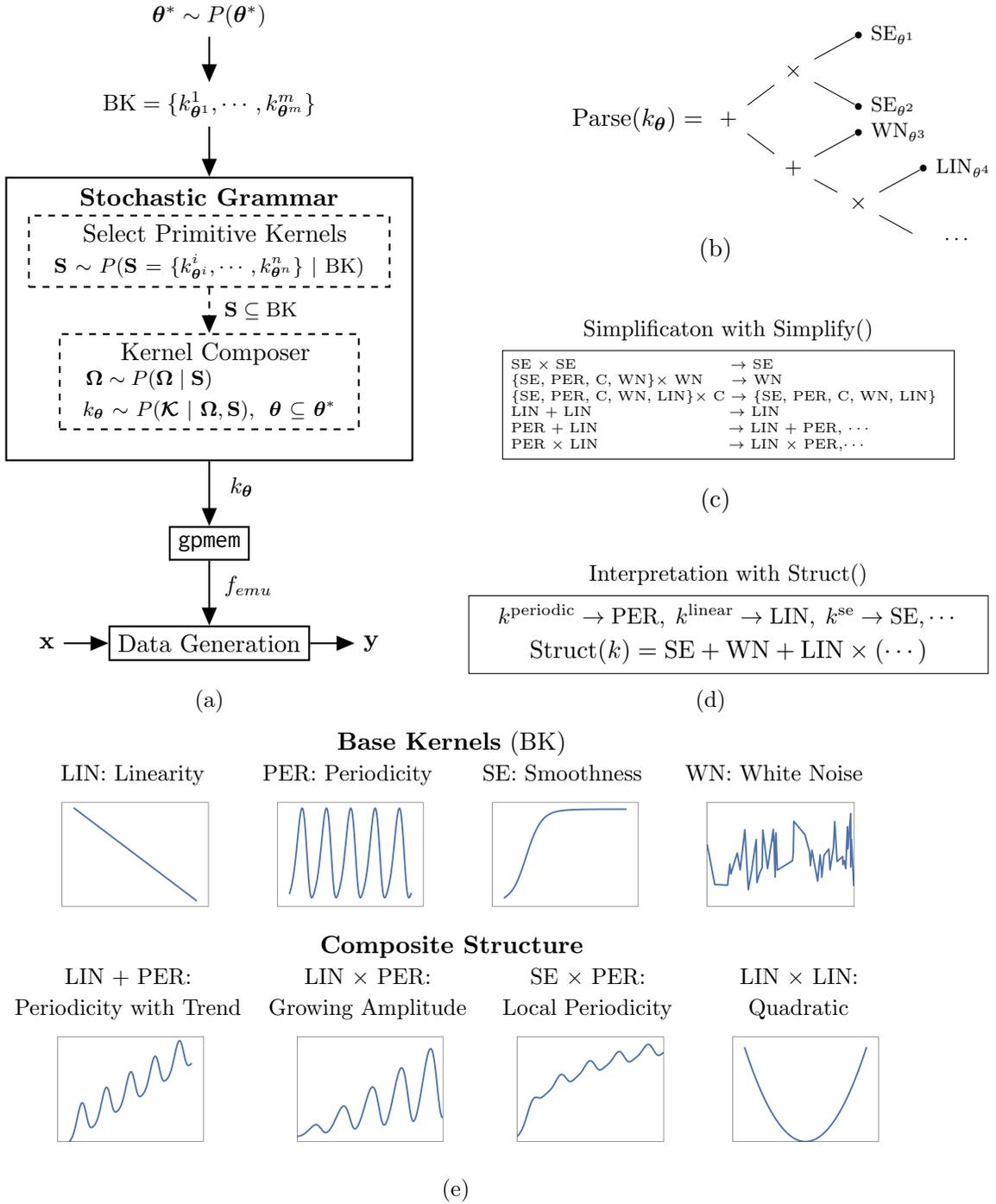}
\caption{\footnotesize (a) Bayesian GP structure learning. A set of
base kernels (BK) with priors on their hyper-parameters serves as hypothesis space
and is supplied as input to the stochastic grammar. The stochastic grammar has
two parts: (i) a sampler that selects a random set $\Sbf$  of primitive kernels from BK
and (ii) a kernel composer that combines the individual base kernels and generates
a composite kernel function
$k_{\thetabf}$. This serves as input for
\gpmem.  We observe value pairs $\xbf,\ybf$ of unstructured time series data on
the bottom of the schematic.
(b) We use $\Parse(\ktheta)$ to parse a structure. (c) kernel functions are
simplified with the $\Simplify()$-operator. The simplifed $\ktheta$ is used as input for
$\Struct()$ (d) which  interprets it symbolically.
Base kernels and compositional kernels are shown in (e) alongside their
interpretation with Struct().}\label{fig:schema}
\end{figure}

With \gpmem\ we can provide a fully Bayesian treatment of this, previously unavaible,
using a stochastic grammar  (see Fig. \ref{fig:schema}).
This allows us to read an unstructured time series and automatically output a high-level,
qualitative description of it. The stochastic grammar takes a set of primitive base kernels 
    $\text{BK}=\{k_{\bm{\theta}^1}^1,\cdots,k_{\bm{\theta}^m}^m\}$

    $\thetabf^*=\{\thetabf^1,\cdots,\thetabf^m\}$ (Fig. \ref{fig:schema} (a) and (b))
We depict the input for the
stochastic grammar in Listing \ref{alg:base_kernels}.

We sample a random subset $\Sbf$ of
the set of supplied base kernels. $\Sbf$ is of size $n \leq m$. We write
\[
\Sbf = \{k_{\bm{\theta}^i}^i,\cdots,k_{\bm{\theta}^n}^n\}
\sim P(\Sbf = \{k_{\bm{\theta}^i}^i,\cdots,k_{\bm{\theta}^n}^n\} \mid
\text{BK}) 
\]
with
\[
P(\Sbf = \{k_{\bm{\theta}^i}^i,\cdots,k_{\bm{\theta}^n}^n\}\mid \text{BK}) =
\frac{n!}{ \mid \Sbf = \{k_{\bm{\theta}^i}^i,\cdots,k_{\bm{\theta}^n}^n\}\mid !}.
\]

BK is assumed to be fixed as the most general margin of our hypothesis space.
In the following, we will drop it in the notation.
The only building block that we are now missing is how to combine the sampled
base kernels into a compositional covariance function (see Fig. \ref{fig:schema}
(b)). For each interaction $i$, we
have to infer whether the data supports a local interaction or a global interaction,
chosing between one out of two algebraic operators
$\bm{\Omega}_i=\{+,\times\}$. The probability for all such decisions is given by a binomial distribution: 
\begin{equation}
P(\bm{\Omega} \mid \Sbf)= {n \choose r}  p_{+\times}^r (1 - p_{+\times})^{n-r}.
\end{equation}
We can write the marginal probability of a kernel function as 
\begin{equation}
P(\Krv \mid \xbf,\ybf,\thetabf ) = \iint \limits_{\bm{\Omega},\Sbf}
P(\Krv \mid \xbf,\ybf,\thetabf,\bm{\Omega},\Sbf) \times P(\bm{\Omega} \mid \Sbf)\times
P(\Sbf)\; \text{\bf d}\bm{\Omega}\, \text{\bf d}\Sbf\,
\end{equation}
with $\bm{\theta}\subseteq \bm{\theta}^*$ as implied by $\Sbf$.
For structure learning with \ac{GP} kernels, a composite kernel is
sampled from $P(\Krv)$ and fed into \gpmem. 
The emulator generated by \gpmem\ observes unstructured time series data.
Venture code for the probabilistic grammar is shown in Listing
\ref{alg:grammar}, code for inference with \gpmem\ in Listing
\ref{alg:structureVent}. 
\begin{mdframed}
\begin{minipage}{\linewidth}
\small
\belowcaptionskip=-10pt
\begin{lstlisting}[mathescape,label=alg:base_kernels,basicstyle=\selectfont\ttfamily,numbers=none,caption={Initialize
Base Kernels BK and $P(n)$},escapechar=\#]
#\linenumber{1}# // Initialize hyper-parameters
#\linenumber{2}#assume theta_1 = tag(scope="hyper-parameters", gamma(5,1));
#\linenumber{3}#assume theta_2 = tag(scope="hyper-parameters", gamma(5,1));
#\linenumber{4}#assume theta_3 = tag(scope="hyper-parameters", gamma(5,1));
#\linenumber{5}#assume theta_4 = tag(scope="hyper-parameters", gamma(5,1));
#\linenumber{6}#assume theta_5 = tag(scope="hyper-parameters", gamma(5,1));
#\linenumber{7}#assume theta_6 = tag(scope="hyper-parameters", gamma(5,1));
#\linenumber{8}#assume theta_7 = tag(scope="hyper-parameters", gamma(5,1));
#\linenumber{9}#
#\linenumber{11}# // Make kernels
#\linenumber{12}#assume lin = apply_function(make_linear, theta_1);
#\linenumber{13}#assume per = apply_function(make_periodic, theta_2, theta_3, theta_4);
#\linenumber{14}#assume se  = apply_function(make_squaredexp, theta_5, theta_6);
#\linenumber{15}#assume wn  = apply_function(make_noise, theta_7);
#\linenumber{16}#
#\linenumber{17}#// Initialize the set of primitive base kernels BK 
#\linenumber{18}#assume BK = list(lin, per, se, wn);
\end{lstlisting}
\end{minipage}
\end{mdframed}

Many equivalent covariance structures can be sampled due to covariance function algebra
and equivalent representations with different parameterization~\citep{lloyd2014automatic}.
To inspect the posterior of these equivalent structures we convert each kernel expression
into a sum of products and subsequently simplify. 
We introduce three different operators that work on kernel functions:
\begin{enumerate}
\item $\Parse(k)$, an operator that parses a covariance function (Fig.
\ref{fig:schema} (b)). 
\item $\Simplify(k)$; this operators simplifies a kernel function $k$ according
to the simplifications that we present in Appendix B and Fig.
\ref{fig:schema} (c).
\item $\Struct(k)$; interprets the structure of a covariance function (Fig.
\ref{fig:schema} (d) and Appendix C), for
example $\Struct(\klin)=\text{LIN}$; it translates the functional structure
into a symbolic expression. 
\end{enumerate}
All base kernels relevant for this work can be found in Appendix A.
\begin{mdframed}
\begin{minipage}{\linewidth}
\small
\belowcaptionskip=-10pt
\begin{lstlisting}[mathescape,label=alg:grammar,basicstyle=\selectfont\ttfamily,numbers=none,caption={
Stochastic Grammar},escapechar=\#]
#\linenumber{1}#// Select a random subset of a set of possible primitive kernels (BK)
#\linenumber{2}#assume select_primitive_kernels = proc(l) {
#\linenumber{3}#  if is_null(l) {
#\linenumber{4}#    l
#\linenumber{5}#  } else {
#\linenumber{6}#    if bernoulli() {
#\linenumber{7}#      pair(first(l), select_primitive_kernels(rest(l)))
#\linenumber{8}#    } else {
#\linenumber{9}#      select_primitive_kernels(rest(l))
#\linenumber{10}#    }
#\linenumber{11}#  }
#\linenumber{12}#};
#\linenumber{13}#// Construct the kernel composition with a composer procedure
#\linenumber{14}#assume kernel_composer = proc(l) {
#\linenumber{15}#  if (size(l) <= 1) {
#\linenumber{16}#    first(l)
#\linenumber{17}#  } else {
#\linenumber{18}#       if (bernoulli()) {
#\linenumber{19}#            add_funcs(first(l),  kernel_composer(rest(l)))
#\linenumber{20}#       } else {
#\linenumber{21}#            mult_funcs(first(l), kernel_composer(rest(l)))
#\linenumber{22}#    }
#\linenumber{23}#  }
#\linenumber{24}#};
#\linenumber{25}#// Select the set primitive kernels that will form the structure
#\linenumber{26}#assume primitive_kernel_selection = tag(scope="grammar", 
#\linenumber{27}#		              permute(select_primitive_kernels(BK)));
#\linenumber{28}#// Compose the structure
#\linenumber{29}#assume K = tag(scope="grammar", 
#\linenumber{30}#        	kernel_composer(primitive_kernel_selection));
\end{lstlisting}

\end{minipage}
\end{mdframed}

\begin{mdframed}
\begin{minipage}{\linewidth}
\small
\belowcaptionskip=-10pt
\begin{lstlisting}[mathescape,label=alg:structureVent,basicstyle=\selectfont\ttfamily,numbers=none,caption={\gpmem\
inference for structure
learning: },escapechar=\#]
#\linenumber{1}#// Apply gpmem 
#\linenumber{2}#assume (f_compute, f_emu) = gpmem(f_look_up, K);
#\linenumber{3}#// Probe all data points
#\linenumber{4}#for (n = 0; n < size(data); n++) { 
#\linenumber{5}#	predict f_compute(first(lookup(data, n)))};
#\linenumber{6}#// Perform inference
#\linenumber{7}#infer repeat(200, do(
#\linenumber{8}#	mh(scope="grammar", steps=1),
#\linenumber{9}#	mh(scope="hyper-parameters", steps=2)));
\end{lstlisting}

\end{minipage}
\end{mdframed}

\newpage
We defined a simple space of covariance structures in a way that allows us to produce results coherent with 
work presented in Automatic Statistician~\citep{duvenaud2013structure,lloyd2014automatic}. The results are illustrated with two data sets.

\myparagraph{Mauna Loa  CO$_2$ data}
\begin{figure}
\centering
\input{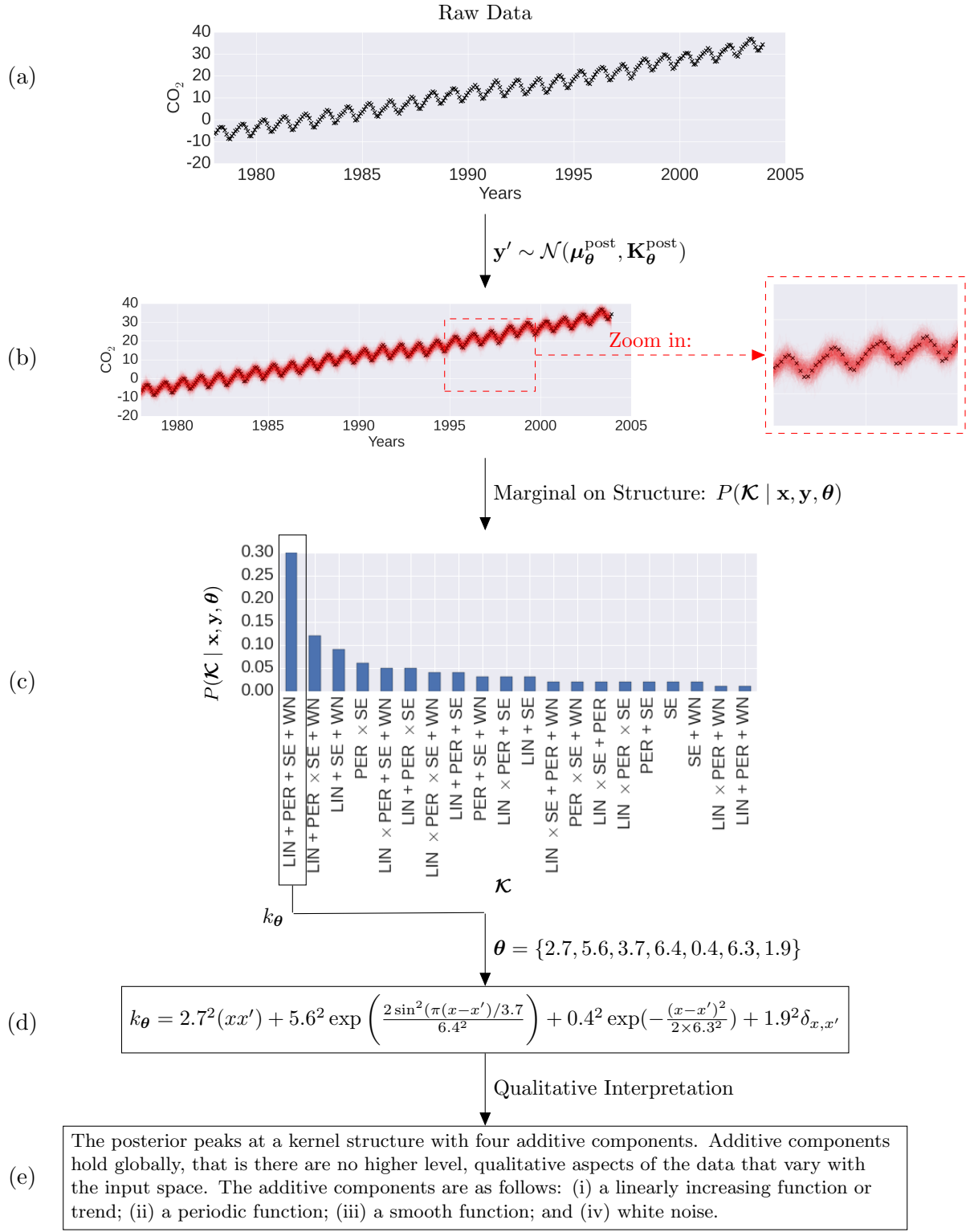}
\caption{\footnotesize Structure Learning. Starting with raw data (a), we fit a \ac{GP}
(b) and compute the posterior distribution on structures (c). We take a sample
of the peak of this distribution ($\text{LIN}+\text{PER}+\text{SE}+\text{WN}$)
including its parameters and write it in functional form (d). We depict the
human readable interpretation (e). We used (d) to plot (b).}\label{fig:posterior}
\end{figure}
We illustrate results in Fig \ref{fig:posterior}. In Fig \ref{fig:posterior} (a) we depict the raw data. 
We see mean centered CO$_2$ measurements of the Mauna Loa Observatory, an atmospheric
baseline station on Mauna Loa, on the island of Hawaii. 
A description of the data set  can be found in  \citealp[][chapter 5]{rasmussen2006gaussian}.  
We use those raw data to compute a posterior on structure, parameters and \ac{GP}
samples.
The latter are shown in  Fig \ref{fig:posterior} (b)
where we zoom in to show how the posterior captures the error bars
adequately.
This posterior of the \ac{GP} is generated with a random sample from the parameters
of the peak of the distribution on structure (Fig \ref{fig:posterior} (c)).
We differentiate between a posterior distribution of kernel functions and  a
posterior distribution of
symbolic expressions describing different kernel structures. 
This allows us to compute the posterior of symbollically equivalent structures,
such as $\Struct(\klin + \kper)=\Struct(\kper + \klin)$. Both structures yield
and addition of a linear kernel and a periodic kernel, that is LIN + PER.
Therefore, we parse $k$ with $\Parse(k)$, we simplify an expression with $\Simplify(k)$ and
then compute $\Struct(k)$.
For the Mauna Loa Co$_2$ data, this distribution peaks at:
\begin{equation}
\Struct(\Ksrv=k)=\text{LIN} + \text{PER} + \text{SE} + \text{WN}.
\end{equation}
We write this kernel equation out in Fig \ref{fig:posterior} (d).
This kernel structure has a natural language interpretation that we spell out in
Fig \ref{fig:posterior} (e), explaining that 
the posterior peaks at a kernel structure with four additive components.
Each of which holds globally, that is there are no higher level, qualitative aspects
of the data that vary with the input space. The additive components for this result are as follows:
\begin{itemize}
\item a linearly increasing function or trend; 
\item a periodic function;
\item a smooth function; and
\item white noise.
\end{itemize}

Previous work on automated kernel discovery~\citep{duvenaud2013structure} illustrated the Mauna Loa data using an RQ kernel.
We resort to the white noise kernel instead of RQ (similar to \citep{lloyd2014automatic}).

\myparagraph{Airline Data}
The second data set (Fig. \ref{fig:posterior_airline}) we depict results for is  the airline 
data set describing monthly totals of international airline passengers (\citealp{box2011time}, according to \citealp{duvenaud2013structure}). 
\begin{figure}
\centering
\input{figs/structure_posterior_airline.tex}
\caption{\footnotesize Structure Learning. Starting with raw data (a), we fit a \ac{GP}
(b) and compute the posterior distribution on structures (c). We take a sample
of the peak of this distribution ($\text{LIN}+\text{PER} \times \text{SE}+\text{WN}$)
including its parameters and write it in functional form (d). We depict the
human readable interpretation (e). We used (d) to plot (b).}\label{fig:posterior_airline}
\end{figure}

We illustrate results for this data set in Fig \ref{fig:posterior_airline}. In Fig \ref{fig:posterior_airline} (a) we depict the raw data. 
Again, the data is mean centered and we use it to 
compute a posterior on structure, parameters and \ac{GP}
samples.
The latter are shown in  Fig \ref{fig:posterior_airline} (b).
This posterior of the \ac{GP} is generated with a random sample from the parameters
of the peak of the distribution on structure (Fig \ref{fig:posterior_airline} (c)).
The posterior over symbolic kernel expressions peaks at:
\begin{equation}
\Struct(\Ksrv = k )=\text{LIN} +  \text{SE} \times \text{PER}+ \text{WN}.
\end{equation}
We write this Kernel equation out in Fig \ref{fig:posterior_airline} (d).
This kernel structure has a natural language interpretation that we spell out in
Fig \ref{fig:posterior_airline} (e), explaining that 
the posterior peaks at a kernel structure with three additive components.
Additive components hold globally, that is there are no higher level, qualitative aspects
of the data that vary with the input space.
The additive components are as follows: 
\begin{itemize}
\item a linearly increasing function or trend;
\item an approximate periodic function; and
\item  white noise.
\end{itemize}
Both datasets served as illustrations in the Automatic Statistician project.

\myparagraph{Querying time series}
With our Bayesian approach to structure learning we can gain valuable insights
into time series data that were previously unavailable.
This is due to our ability to estimate posterior marginal probabilities over the kernel structure.
Over this marginal, we define boolean search operations that allow us to query the data
for the probability of certain structures to hold true globally.
\begin{align}
\label{eq:bool_present}
P(\Struct(\Ksrv=k) \mid \xbf,\ybf,\thetabf) =& \frac{1}{T}
\sum\limits_{t=1}^T \Cont(k,k^t)\\
&\text{where}\, \Cont(k,k^t) = \begin{cases}
  1, & \text{if } k \underset{global}{\in} k^t, \\
  0, & \text{otherwise}.
\end{cases} 
\end{align}
to ask whether it is true that a global structure $\Ksrv=k$ is present. $T$
is the number of all posterior samples for $\Ksrv$ and $k^t$ is one such
sample. $\underset{global}{\in} k^t$ reads as ``is one of $k^t$'s global
functional components".
We can now ask simple questions, for example:
\begin{quotation}
Is there white noise in the data?
\end{quotation}
To answer this question we set $\Struct(\Ksrv) = $WN in (\ref{eq:bool_present}). We write this in
shorthand as $\WNK$. Similarly, we write $\LINK$, $\PERK$ and $\SEK$. 
We can also formulate more sophisticated search operations using Boolean operators such as AND ($\land$) and OR ($\lor$).
The AND operator is defined as follows:
\[
P(\Struct(\Ksrv^a=k^a)\, \land\, \Struct(\Ksrv^b=k^b) \mid \xbf,\ybf, \thetabf)  = \frac{1}{N}
\sum\limits_{n=1}^N \Cont(k^a,k^t)\, \land \, \Cont(k^b,k^t)\;\;
\]
where
\[
\Cont(k^a,k^t) \land \Cont(k^b,k^t) = \begin{cases}
  1, & \text{if } k^a\, \text{and}\, k^b  \underset{global}{\in} k^t, \\
  0, & \text{otherwise}\end{cases}.
\]

\begin{figure}
\centering
\input{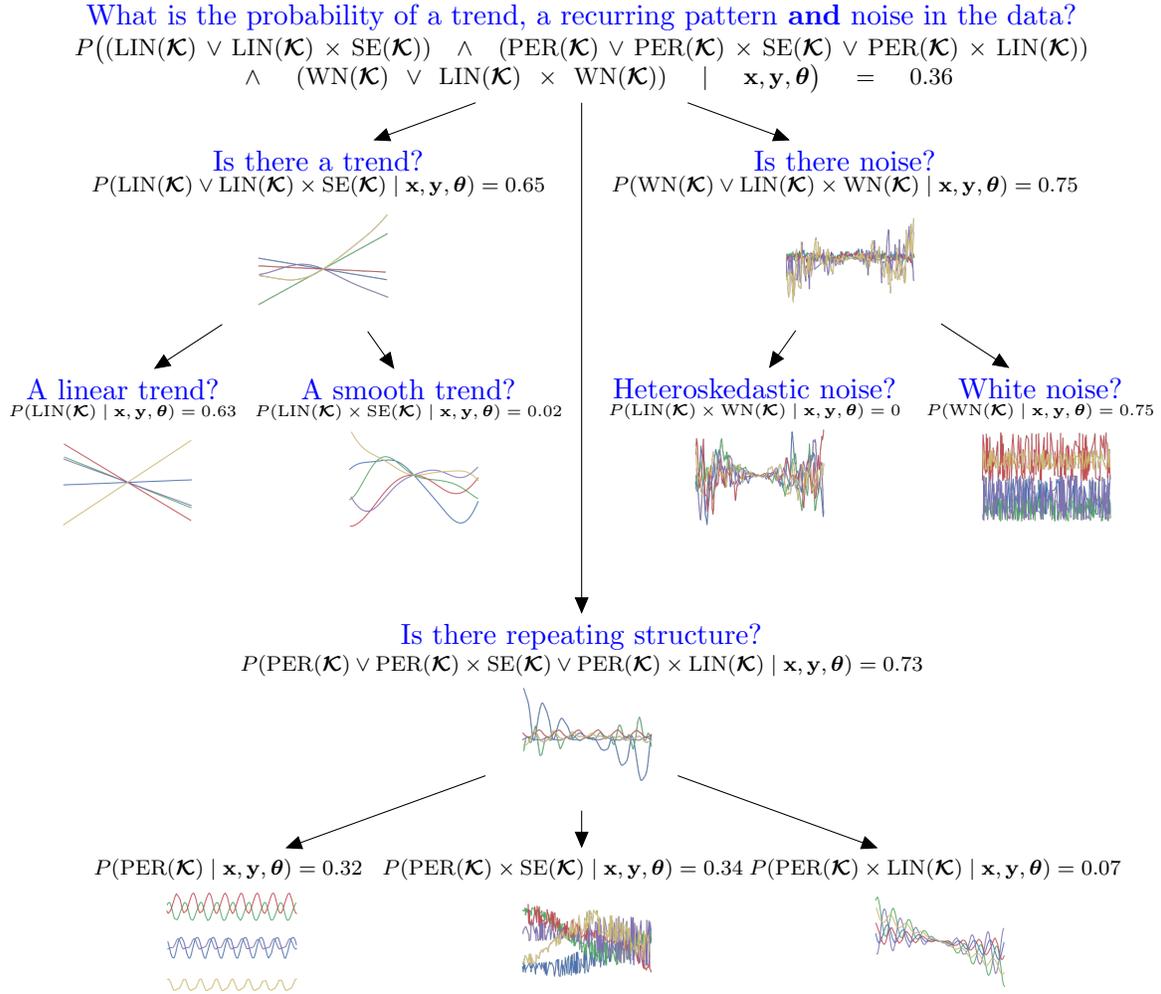}
\caption{{\footnotesize \bf Querying structural motifs in in time series using posterior inference
over kernel structure.} The kernel structure serves as a way to formulate
natural language questions about the data (blue). The initial question of interest
(top) is a fairly
general one: ``What is the probability of a trend, a recurring
pattern and noise in the data?" Below the natural language version of this
question, the same question is formulated as an inference problem (black) over the
marginal probability on kernels with Boolean operators AND ($\land$) and OR ($\lor$). 
To gain  a deeper understanding of specific motifs in the time series more specific queries can
be written.
On the right, a query asks whether there is noise in the data (blue) by computing the disjunction of the marginal
of a global white noise kernel and a multiplication between a linear and a white
noise kernel (black). Samples from the predictive prior $\ybf_*$ of such kernels give an
indication of the qualitative aspects that a kernel structure implies (coloured curves below
the marginal). 
If the probability that there is noise in the data is high, it makes sense
to drill even deeper asking more detailed questions. With regards to noise, this
translates to querying whether or not the data supports the hypothesis that there is
heteroskedastic noise or white noise. Queries for motifs of repeating structure
are shown in the middle of the tree, queries related to trends on the left.}\label{fig:query}
\end{figure}
By estimating $P(\LINK \land \WNK \mid \xbf, \ybf, \thetabf)$ we can use this operator to ask questions such as 
\begin{quotation}
Is there a linear component AND a white noise in the data? 
\end{quotation}
Finally, we define the logical OR as
\begin{align*}
P(\Struct(\Ksrv^a)\, \lor\, \Struct(\Ksrv^b) \mid \xbf, \ybf, \thetabf)
=& P(\Struct(\Ksrv^a) \mid \xbf, \ybf, \thetabf) + P(\Struct(\Ksrv^b) \mid \xbf, \ybf, \thetabf)\\
 &- P(\Struct(\Ksrv^a) \land \Struct(\Ksrv^b) \mid \xbf, \ybf, \thetabf)
\end{align*}
where we drop $=k$ for readability. This allows us to ask questions about structures that are logically connected with OR, such as:
\begin{quotation}
Is there white noise or heteroskedastic noise?
\end{quotation}
by estimating $P(\LINK \times \WNK\;\;{\large\lor}\;\; \WNK \mid
\xbf, \ybf, \thetabf)$.
We know that noise can either be heteroskedastic or white,
and we also know due to simple manipulations using kernel algebra
that  $\text{LIN} \times \WN$ and $\text{WN}$ are the only possible ways to
construct noise with kernel composition. This allows us to generalize the 
question above to:
\begin{quotation}
Is there noise in the data? 
\end{quotation}
where we write the marginal posterior on qualitative structure for noise:
\begin{equation}
P\big(\Struct(\Ksrv)=\text{Noise} \mid \xbf, \ybf, \thetabf\big) = P(\LINK \times
\WNK\;\;{\large\lor}\;\; \WNK \mid \xbf, \ybf, \thetabf).
\end{equation}
From a methodological perspective, this allows us to start with general queries and 
subsequently formulate follow up queries that go into more detail.
For example, we could start with a general query, such as:
\begin{quotation}
What is the probability of a trend, a recurring pattern {\bf and} noise in the data?
\end{quotation}
and then follow up with more detailed questions (Fig \ref{fig:query}).

This way of querying data for their statistical implications is in stark contrast to what previous research in automatic kernel construction was able to provide.
We could view our approach as a time series search engine which allows us to test whether or not certain structures can be found
in an available time series.
Another way to view this approach is as a new language to interact with the world.
Real-world observations often come with time-stamps and in form
of continuous valued sensor measurements.  
We provide the toolbox to query such observations in a similar manner as
one would query a knowledge base in a logic programming language.

\subsection{Bayesian optimization}

\label{sec:thompson}

The final application demonstrating the power of \gpmem\ illustrates its use in Bayesian optimization. We introduce Thompson sampling, the basic solution strategy
underlying the Bayesian optimization with \gpmem.
Thompson sampling~\cite{thompson1933likelihood} is a widely used Bayesian
framework for addressing the trade-off between exploration and exploitation in
multi-armed (or continuum-armed) bandit problems.  
We cast the multi-armed bandit problem as a one-state Markov
decision process (MDP), and describe how Thompson sampling can be used to choose
actions for that MDP.

The MDP can be described as follows: An agent is to take a sequence of actions $x_1, x_2,
\ldots$ from a (possibly infinite) set of possible actions $\Acal$.  After each
action, a reward $y \in \R$ is received, according to an unknown conditional
distribution $P_\true\pn{y \mvert x}$.  The agent's goal is to maximize the
total reward received for all actions in an online manner.  In Thompson
sampling, the agent accomplishes this by placing a prior distribution
$P(\ttheta)$ on the possible ``contexts'' $\ttheta \in \Theta$.  Here a context
is a believed model of the conditional distributions $\{P\pn{x \mvert y}\}_{x
\in \Acal}$, or at least, a believed statistic of these conditional
distributions which is sufficient for deciding an action $x$.  If actions are
chosen so as to maximize expected reward, then one such sufficient statistic is
the believed conditional mean $V \pn{x \mvert \ttheta} = \Ebkt{y \mvert
x;\ttheta}$, which can be viewed as a believed value function.  For
consistency with what follows, we will assume our context $\ttheta$ takes the
form $(\thetabf, \abf, \rbf)$ where $\abf$ is the vector of past
actions, $\rbf$ is the vector of their rewards, and $\bm{\theta}$ (the
``semicontext'') contains any other information that is included in the context.

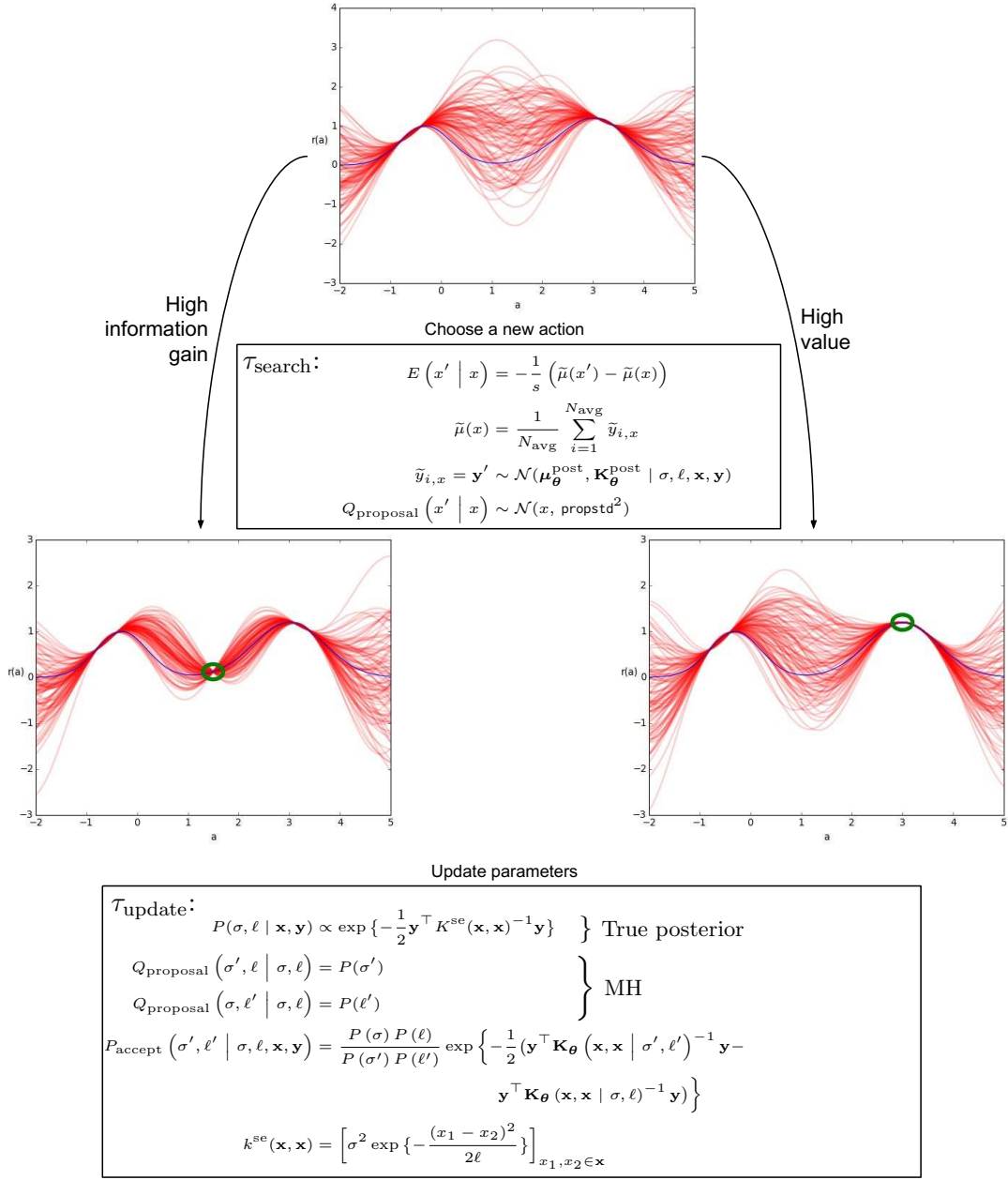
\begin{figure}
  \centering
\begin{overpic}[width=1.05\textwidth]{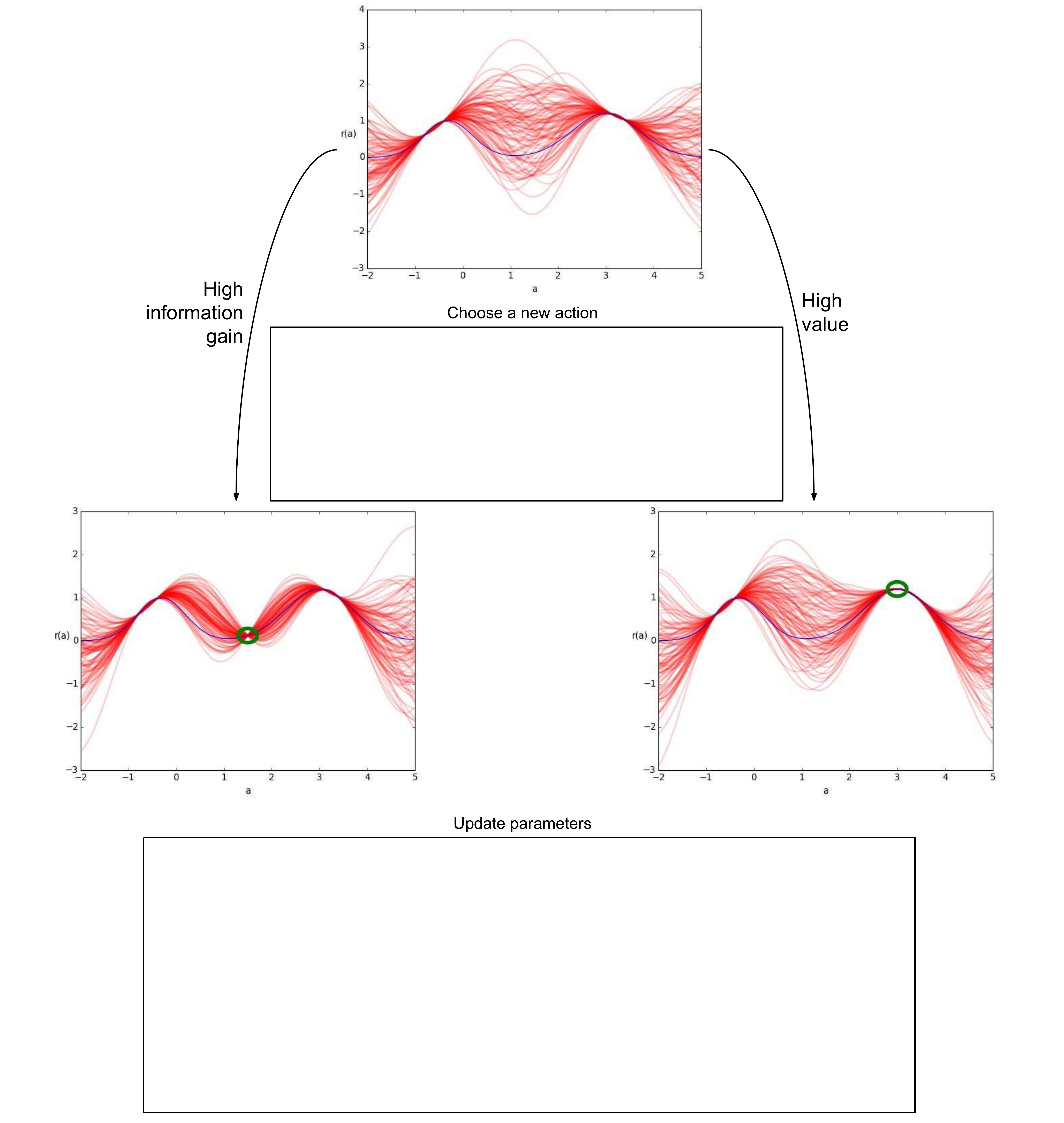}
\put(150,300){{\parbox{0.4\linewidth}{%
    \tiny
    \begin{align*}
	E\pn{x' \mvert x} &= -\frac1s \pn{\widetilde{\mu}(x') -
	\widetilde{\mu}(x)} \\
	\widetilde{\mu}(x) &= \frac{1}{N_\avg} \sum_{i=1}^{N_\avg}
	\widetilde{y}_{i,x}\\ 
	\widetilde{y}_{i,x}=\yprime &\sim \Ncal(\mupost,  \Kpost \mid
\sigma,\ell,\abf,\rbf)\\ 
	Q_\proposal\pn{x' \mvert x} &\sim \Ncal(x,\,\propstd^2)
    \end{align*}}}}
\put(118,331){$\tau_\search$:}
\put(62,58){{\parbox{0.4\linewidth}{%
    \tiny
    \begin{align*}
  P(\sigma,\ell \mid \abf,\rbf) &\propto \exp \big \{ {-\frac{1}{2}}\rbf^\top
\Kse(\abf,\abf)^{-1}\rbf \big \}\\
  Q_\proposal\pn{\sigma',\ell \mvert \sigma,\ell} &= P(\sigma')\\
  Q_\proposal\pn{\sigma,\ell' \mvert \sigma,\ell} &= P(\ell')\\
  P_\accept\pn{\sigma',\ell' \mvert \sigma,\ell,\abf,\rbf}
  &=
   \frac{
    P\pn{\sigma}P\pn{\ell}
    }{
    P\pn{\sigma'}P\pn{\ell'}
    } 
   \exp\bigg \{ {-\frac12}\big(
    \rbf^\top \Ktheta\pn{\abf, \abf \mvert \sigma', \ell'}^{{-1}} \rbf -\\ 
   &\;\;\;\;\;\;\;\; \;\;\;\;\;\;\;\;  \;\;\;\;\;\;\;\;  \;\;\;\;\;\;\;\;  \;\;\;   \rbf^\top \Ktheta\pn{\abf, \abf \mvert \sigma, \ell}^{{-1}} \rbf
  \big)\bigg \}\\
   \kse(\abf,\abf)&=\bigg[\sigma^2 \exp \big \{{-\frac{(x_1 -
x_2)^2}{2\ell}}\big \}\bigg]_{x_1,x_2 \in \abf}
    \end{align*}}}}
\put(65,112){$\tau_\update$:}
\put(252,78){\bigg \} \footnotesize MH}
\put(250,102){ \} \footnotesize True posterior }
\end{overpic}
  \caption{\footnotesize
    Two possible actions (in green) for an iteration of Thompson sampling.  The
    believed distribution on the value function $V$ is depicted in red.  In this
    example, the true reward function is deterministic, and is drawn in blue.
    The action on the right receives a high reward, while the action on the left
    receives a low reward but greatly improves the accuracy of the believed
    distribution on $V$.  The transition operators $\tau_\search$ and
    $\tau_\update$ are described in Section \ref{sec:math-spec}.
  }
  \label{fig:slide2}
\end{figure}

In this setup, Thompson sampling has the following steps:
\begin{algorithm}[H]
  \singlespacing
  Repeat as long as desired:
  \begin{enumerate}
    \item\label{itm:thompson-step-sample} {\bf Sample.} Sample a semicontext
      $\thetabf \sim P(\thetabf)$.
    \item\label{itm:thompson-step-search} {\bf Search (and act).} Choose an
      action $x \in \Acal$ which (approximately) maximizes $V\pn{x \mvert
      \ttheta} = \Ebkt{y \mvert x; \ttheta} = \Ebkt{y \mvert x;\,\thetabf,
      \abf, \rbf}$.
    \item {\bf Update.} Let $y_\true$ be the reward received for action $x$.
      Update the believed distribution on $\thetabf$, i.e., $P(\thetabf) \gets
      P_\rmnew(\thetabf)$ where $P_\rmnew(\thetabf) = P\pn{\thetabf \mvert x \mapsto
      y_\true}$.
  \end{enumerate}
  \caption{Thompson sampling.}
  \label{alg:thompson}
\end{algorithm}
Note that when $\Ebkt{y|x;\ttheta}$ (under the sampled value of $\thetabf$ for
some points $x$) is far from the true value $\Ebkt[P_\true]{y \mvert x}$, the
chosen action $x$ may be far from optimal, but the information gained by probing
action $x$ will improve the belief $\ttheta$.  This amounts to ``exploration.''
When $\Ebkt{y \mvert x;\ttheta}$ is close to the true value except at points $x$
for which $\Ebkt{y \mvert x;\ttheta}$ is low, exploration will be less likely to
occur, but the chosen actions $x$ will tend to receive high rewards.  This
amounts to ``exploitation.'' The trade-off between exploration and exploitation
is illustrated in Figure \ref{fig:slide2}.  Roughly speaking, exploration will
happen until the context $\ttheta$ is reasonably sure that the unexplored
actions are probably not optimal, at which time the Thompson sampler will
exploit by choosing actions in regions it knows to have high value.

Typically, when Thompson sampling is implemented, the search over contexts
$\ttheta \in \Theta$ is limited by the choice of representation.  In
traditional programming environments, $\thetabf$ often consists of a few
numerical parameters for a family of distributions of a fixed functional
form.  With work, a mixture of a few functional forms is possible; but
without probabilistic programming machinery, implementing a rich context
space $\Theta$ would be an unworkably large technical burden.  In a
probabilistic programing language, however, the representation of
heterogeneously structured or infinite-dimensional context spaces is quite
natural.  Any computable model of the conditional distributions
$\br{P\pn{y \mvert x}}_{x \in \Acal}$ can be represented as a stochastic
procedure $(\lambda (x) \ldots)$.  Thus, for computational Thompson sampling,
the most general context space $\widehat\Theta$ is the space of program texts.
Any other context space $\Theta$ has a natural embedding as a subset of
$\widehat\Theta$.
\myparagraph{A Mathematical Specification}\label{sec:math-spec}
We now describe a particular case of Thompson sampling with the following properties:
\begin{itemize}
  \item The regression function has a Gaussian process prior.
  \item The actions $x_1,x_2,\ldots \in \Acal$ are chosen by a Metropolis-like search
    strategy with Gaussian drift proposals.
  \item The hyperparameters of the Gaussian process are inferred using
    Metropolis--Hastings sampling after each action.
\end{itemize}

In this version of Thompson sampling, the contexts $\ttheta$ are Gaussian
processes over the action space $\Acal = [-20, 20] \subseteq \R$.  That is,
\[ V \sim \mathcal{GP}(\mu,k), \]
where the mean $\mu$ is a computable function $\Acal \to \R$ and the covariance
$k$ is a computable (symmetric, positive-semidefinite) function $\Acal \times
\Acal \to \R$.  This represents a Gaussian process $\br{R_a}_{a \in \Acal}$,
where $R_x$ represents the reward for action $x$. We write past actions as
$\abf$ and past rewards as $\rbf$. Computationally, we represent
a context as a data structure
\[ \ttheta = (\thetabf, \abf, \rbf) = (\mu, k,\thetabf, \abf, \rbf), \]
where $\mubf$ is a procedure to be used as the prior mean function and
$k$ is a procedure to be used as the prior covariance function, parameterized by 
$\thetabf$.
As above set $\mu \equiv 0$.

Note that the context space $\Theta$ is not a finite-dimensional parametric
family, since the vectors $\abf$ and $\rbf$ grow as more samples are
taken.  $\Theta$ is, however, representable as a computational
procedure together with parameters and past samples, as we do in the
representation $\ttheta = (\mu, k, \thetabf, \abf, \rbf)$.

We combine the Update and Sample steps of Algorithm \ref{alg:thompson} by
running a Metropolis--Hastings (MH) sampler whose stationary distribution is the
posterior $P\pn{\thetabf \mvert \abf, \rbf}$.  The functional forms of
$\mu$ and $k$ are fixed in our case, so inference is only done
over the parameters $\thetabf = \br{\sigma,\ell}$; hence we equivalently write
$P\pn{\sigma,\ell \mvert \abf, \rbf}$ for the stationary
distribution.  We make MH proposals to one variable at a time, using the prior
as proposal distribution:
\[
  Q_\proposal\pn{\sigma',\ell \mvert \sigma,\ell} = P(\sigma')
\]
and
\[
  Q_\proposal\pn{\sigma,\ell' \mvert \sigma,\ell} = P(\ell').
\]
The MH acceptance probability for such a proposal is
\[
  P_\accept\pn{\sigma',\ell' \mvert \sigma,\ell}
  =
  \min\br{1,\ \frac{
    Q_\proposal\pn{\sigma,\ell \mvert \sigma',\ell'}
    }{
    Q_\proposal\pn{\sigma',\ell' \mvert \sigma,\ell}
    }
  \cdot
  \frac{
    P\pn{\abf,\rbf \mvert \sigma',\ell'}
    }{
    P\pn{\abf,\rbf \mvert \sigma,\ell}
    }}
\]
Because the priors on $\sigma$ and $\ell$ are uniform in our case, the term
involving $Q_\proposal$ equals $1$ and we have simply
\begin{align*}
  P_\accept\pn{\sigma',\ell' \mvert \sigma,\ell}
  &=
  \min\br{1,\ \frac{
    P\pn{\abf,\rbf \mvert \sigma',\ell'}
    }{
    P\pn{\abf,\rbf \mvert \sigma,\ell}
    }} \\[2mm]
  &=
  \min\bigg\{1,\ \exp\bigg( -\frac12\bigg(
    \rbf^T K \pn{\abf, \abf \mvert \sigma', \ell'}^{-1} \rbf \\
  & \qquad\qquad\qquad\qquad\qquad -
    \rbf^T K \pn{\abf, \abf \mvert \sigma, \ell}^{-1} \rbf
  \bigg)\bigg)\bigg\}.
\end{align*}
The proposal and acceptance/rejection process described above define a
transition operator $\tau_\update$ which is iterated a specified number of
times; the resulting state of the MH Markov chain is taken as the sampled
semicontext $\thetabf$ in Step \ref{itm:thompson-step-sample} of Algorithm
\ref{alg:thompson}.

For Step \ref{itm:thompson-step-search} (Search) of Thompson sampling, we
explore the action space using an MH-like transition operator $\tau_\search$.
As in MH, each iteration of $\tau_\search$ produces a proposal which is either
accepted or rejected, and the state of this Markov chain after a specified
number of steps is the new action $x$.  The Markov chain's initial state is the
most recent action, and the proposal distribution is Gaussian drift:
\[ Q_\proposal\pn{x' \mvert x} \sim \Ncal(x,\,\propstd^2), \]
where the drift width $\propstd$ is specified ahead of time.  The acceptance
probability of such a proposal is
\[ P_\accept\pn{x' \mvert x} = \min\br{1,\ \exp\pn{-E\pn{x' \mvert x}}}, \]
where the energy function $E\pn{\bullet \mvert a}$ is given by a Monte Carlo
estimate of the difference in value from the current action:
\[ E\pn{x' \mvert x} = -\frac1s \pn{\widetilde{\mu}(x') - \widetilde{\mu}(x)} \]
where
\[ \widetilde{\mu}(x) = \frac{1}{N_\avg} \sum_{i=1}^{N_\avg} \widetilde{y}_{i,x} \]
and
\[ \widetilde{y}_{i,x}=\yprime \sim \Ncal(\mupost, \Kpost) \]
and $\br{\widetilde{y}_{i,x}}_{i=1}^{N_\avg}$ are i.i.d.\ for a fixed $x$.
(In the above, $\mupost$ and $\Kpost$ are the mean and variance of a posterior
sample at the single point $\xprime = (x^\prime)$.)
Here the temperature parameter $s \geq 0$ and the population size $N_\avg$ are
specified ahead of time.  Proposals of estimated value higher than that of the current action are
always accepted, while proposals of estimated value lower than that of the
current action are accepted with a probability that decays exponentially
with respect to the difference in value.
The rate of the decay is determined by the temperature parameter $s$,
where high temperature corresponds to generous acceptance probabilities.
For $s=0$, all proposals of lower value are rejected; for $s=\infty$, all
proposals are accepted.
For points $x$ at which the posterior mean $\mupost$ is low but the
posterior variance $\Kpost$ is high, it is possible (especially when
$N_\avg$ is small) to draw a ``wild'' value of $\widetilde{\mu}(x)$, resulting in a
favorable acceptance probability.

Indeed, taking an action $x$ with low estimated value but high uncertainty
serves the useful function of improving the accuracy of the estimated value
function at points near $x$ (see Figure \ref{fig:slide2}).\footnote{
  At least, this is true when we use a smoothing prior covariance function such
  as the squared exponential.
}$^,$\footnote{
  For this reason, we consider the sensitivity of $\muhat$ to uncertainty to be
  a desirable property; indeed, this is why we use $\muhat$ rather than the
  exact posterior mean $\mu$.
}
We see a complete probabilistic program with \gpmem\ implementing Bayesian optimization
with Thompson Sampling and both, uniform proposals and drift proposals below
(Listing \ref{alg:init_bayesopt},\ref{alg:uniform_bayesopt} and \ref{alg:drift_bayesopt}).
 \begin{mdframed}
\begin{minipage}{\linewidth}
\small
\belowcaptionskip=-10pt
\begin{lstlisting}[caption={Initialize \gpmem\ for Bayesian
optimization},mathescape,numbers=none,label=alg:init_bayesopt,escapechar=\#]
#\linenumber{1}#assume sf = tag(scope="hyper", uniform_continuous(0, 10));
#\linenumber{2}#assume l = tag(scope="hyper",  uniform_continuous(0, 10));
#\linenumber{3}#assume se = make_squaredexp(sf, l);
#\linenumber{4}#assume blackbox_f = get_bayesopt_blackbox();
#\linenumber{5}#assume (f_compute, f_emulate) = gpmem(blackbox_f, se);
\end{lstlisting}

\end{minipage}
\end{mdframed}

 \begin{mdframed}
\begin{minipage}{\linewidth}
\small
\belowcaptionskip=-10pt
\begin{lstlisting}[caption={Bayesian optimization with uniformly distributed
proposals},mathescape,numbers=none,label=alg:uniform_bayesopt,escapechar=\#]
#\linenumber{1}#// A naive estimate of the argmax of the given function
#\linenumber{2}#define mc_argmax = proc(func) {
#\linenumber{3}#  candidate_xs = mapv(proc(i) {uniform_continuous(-20, 20)},
#\linenumber{4}#                      arange(20));
#\linenumber{5}#  candidate_ys = mapv(func, candidate_xs);
#\linenumber{6}#  lookup(candidate_xs, argmax_of_array(candidate_ys))
#\linenumber{7}#};
#\linenumber{8}#
#\linenumber{9}#// Shortcut to sample the emulator at a single point without packing
#\linenumber{10}#// and unpacking arrays
#\linenumber{11}#define emulate_pointwise = proc(x) {
#\linenumber{12}#  run(sample(lookup(f_emulate(array(unquote(x))), 0)))
#\linenumber{13}#};
#\linenumber{14}#
#\linenumber{15}#// Main inference loop
#\linenumber{16}#infer repeat(15, do(pass,
#\linenumber{17}#  // Probe V at the point mc_argmax(emulate_pointwise)
#\linenumber{18}#  predict(f_compute(unquote(mc_argmax(emulate_pointwise)))),
#\linenumber{15}#  // Infer hyper-parameters
#\linenumber{20}#  mh(scope="hyper", steps=50)));
\end{lstlisting}

\end{minipage}
\end{mdframed}

 \begin{mdframed}
\begin{minipage}{\linewidth}
\small
\belowcaptionskip=-10pt
\begin{lstlisting}[caption={Bayesian optimization with Gaussian drift
proposals},mathescape,numbers=none,label=alg:drift_bayesopt,escapechar=\#]
#\linenumber{1}#// A naive estimate of the argmax of the given function
#\linenumber{2}#define mc_argmax = proc(func) {
#\linenumber{3}#  candidate_xs = mapv(proc(x) {normal(x, 1)},
#\linenumber{4}#                 fill(20,last));
#\linenumber{5}#  candidate_ys = mapv(func, candidate_xs);
#\linenumber{6}#  lookup(candidate_xs, argmax_of_array(candidate_ys))
#\linenumber{7}#};
#\linenumber{8}#
#\linenumber{9}#// Shortcut to sample the emulator at a single point without packing
#\linenumber{10}#// and unpacking arrays
#\linenumber{11}#define emulate_pointwise = proc(x) {
#\linenumber{12}#  run(sample(lookup(f_emulate(array(unquote(x))), 0)))
#\linenumber{13}#};
#\linenumber{15}#
#\linenumber{16}#// Initialize helper variables
#\linenumber{17}#assume previous_point = uniform_continuous(-20,20);
#\linenumber{18}#run(observe(previous_point ,run(sample(previous_point)),prev));
#\linenumber{19}#
#\linenumber{20}#// Main inference loop
#\linenumber{21}#infer repeat(15, do(pass,
#\linenumber{22}#  // find the next point with mc argmax
#\linenumber{23}#  next_point <- action(mc_argmax(
#\linenumber{24}#		        emu_pointwise,run(sample(previous_point)))),
#\linenumber{25}#  // Probe V at the point mc_argmax(emu_pointwise)
#\linenumber{26}#  predict(first(package)(unquote(next_point))),
#\linenumber{27}#  // Clear the previous point
#\linenumber{28}#  forget(quote(prev)),
#\linenumber{29}#  // Remember the current probe as the previous one for the next iter.
#\linenumber{30}#  observe(previous_point, next_point,prev),
#\linenumber{31}#  // Infer hyper-parameters
#\linenumber{32}#  mh(scope="hyper", steps=50)));
\end{lstlisting}

\end{minipage}
\end{mdframed}

 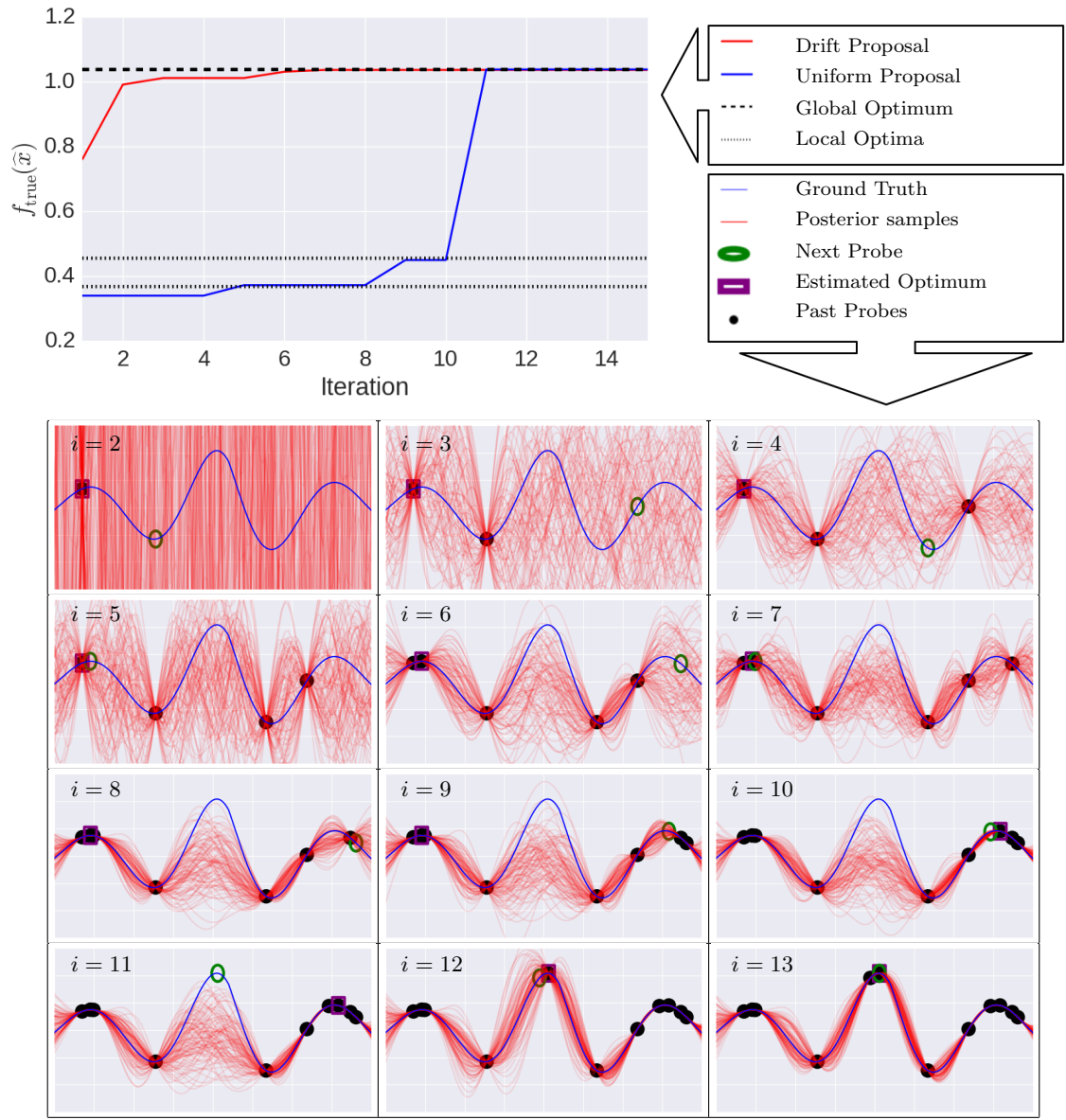
\begin{figure}
  \setlength{\tabcolsep}{1pt} 
 \centering
\begin{tabular}{rl}
\includegraphics[width=0.6\textwidth]{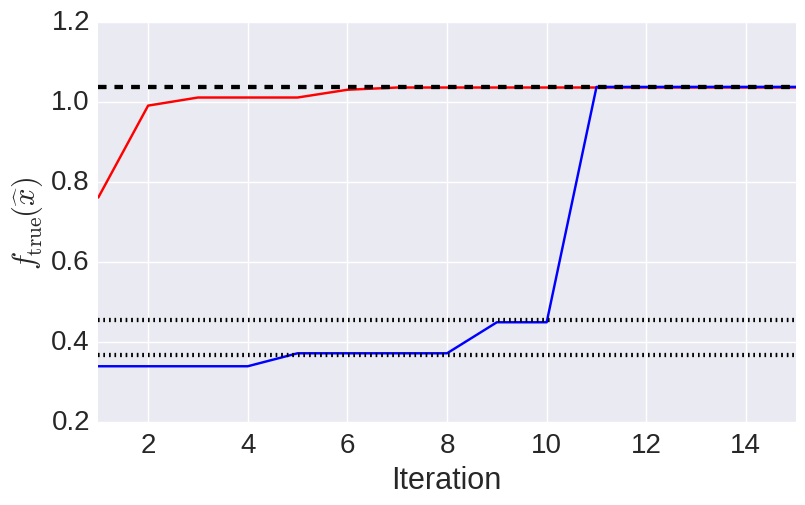}&\includegraphics[width=0.4\textwidth]{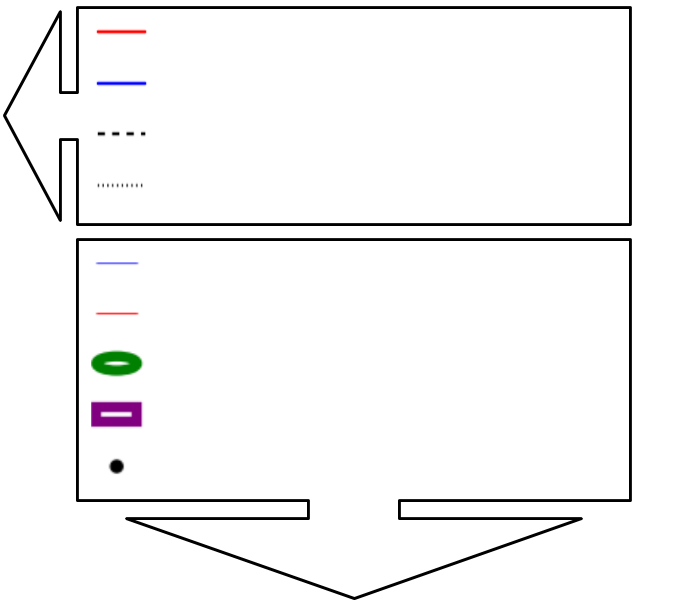}
\put(-119,142){\scriptsize Drift Proposal}
\put(-119,130){\scriptsize Uniform Proposal}
\put(-119,117){\scriptsize Global Optimum}
\put(-119,105){\scriptsize Local Optima}
\put(-119,85){\scriptsize Ground Truth}
\put(-119,73){\scriptsize Posterior samples}
\put(-119,60){\scriptsize Next Probe}
\put(-119,48){\scriptsize Estimated Optimum}
\put(-119,36){\scriptsize Past Probes}
\end{tabular}
\begin{tabular}{llll}\hline 
\multicolumn{1}{|l|}{\raisebox{-0.5\height}{\includegraphics[width=0.3\textwidth]{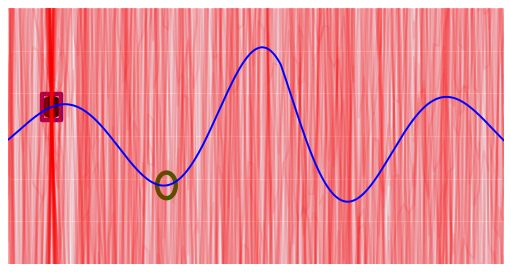}}}&
\multicolumn{1}{l|}{\raisebox{-0.5\height}{\includegraphics[width=0.3\textwidth]{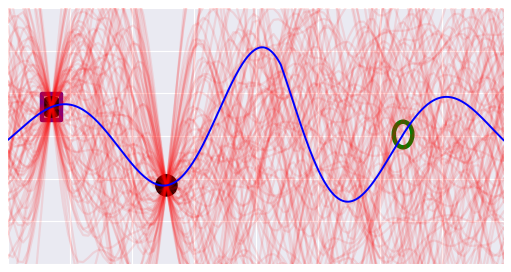}}}&
\multicolumn{1}{l|}{\raisebox{-0.5\height}{\includegraphics[width=0.3\textwidth]{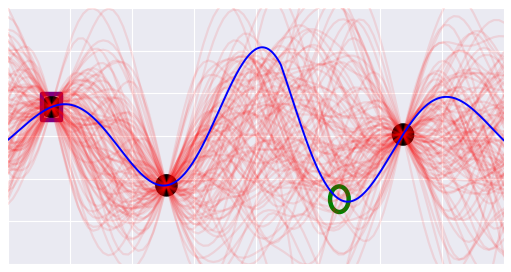}}} \\ \hline
\multicolumn{1}{|l|}{\raisebox{-0.5\height}{\includegraphics[width=0.3\textwidth]{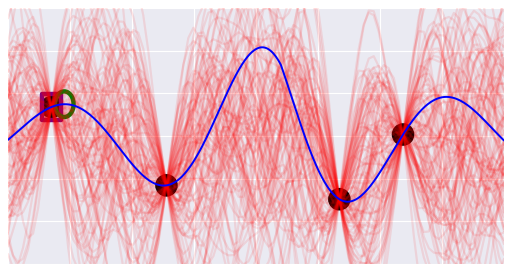}}}&
\multicolumn{1}{l|}{\raisebox{-0.5\height}{\includegraphics[width=0.3\textwidth]{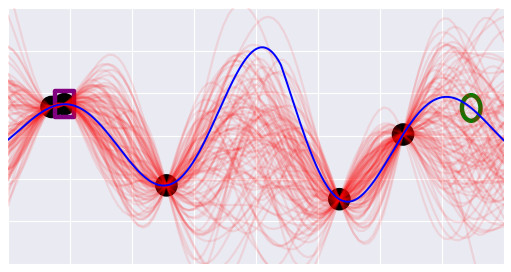}}}&
\multicolumn{1}{l|}{\raisebox{-0.5\height}{\includegraphics[width=0.3\textwidth]{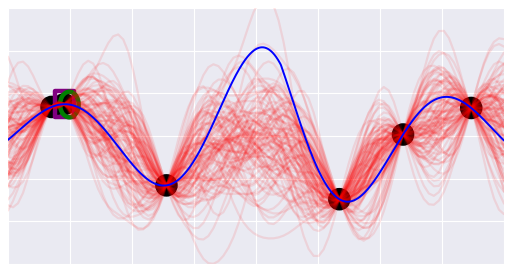}}} \\ \hline
\multicolumn{1}{|l|}{\raisebox{-0.5\height}{\includegraphics[width=0.3\textwidth]{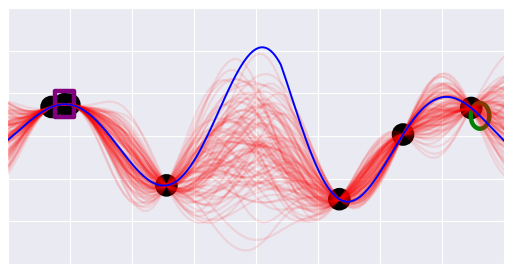}}}&
\multicolumn{1}{l|}{\raisebox{-0.5\height}{\includegraphics[width=0.3\textwidth]{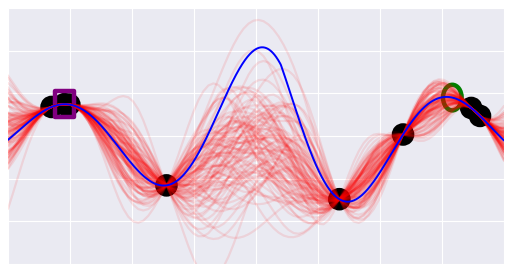}}}&
\multicolumn{1}{l|}{\raisebox{-0.5\height}{\includegraphics[width=0.3\textwidth]{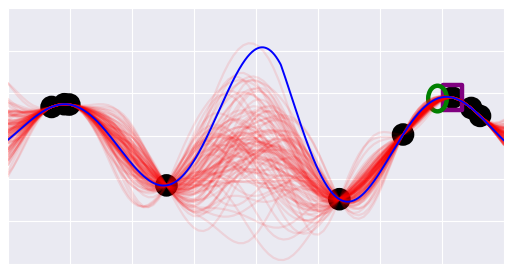}}} \\ \hline
\multicolumn{1}{|l|}{\raisebox{-0.5\height}{\includegraphics[width=0.3\textwidth]{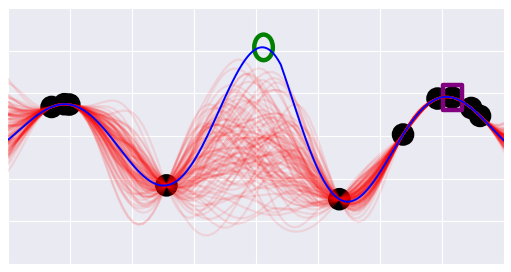}}}&
\multicolumn{1}{l|}{\raisebox{-0.5\height}{\includegraphics[width=0.3\textwidth]{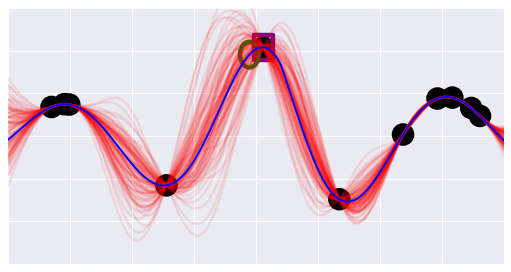}}}&
\multicolumn{1}{l|}{\raisebox{-0.5\height}{\includegraphics[width=0.3\textwidth]{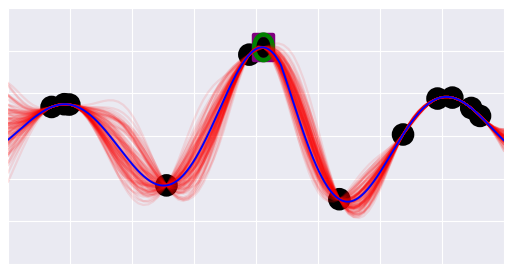}}} \\ \hline
\end{tabular}
\put(-387,130){\footnotesize$i = 2$}
\put(-255,130){\footnotesize$i = 3$}
\put(-123,130){\footnotesize$i = 4$}
\put(-387,62){\footnotesize$i = 5$}
\put(-255,62){\footnotesize$i = 6$}
\put(-123,62){\footnotesize$i = 7$}
\put(-387,-8){\footnotesize$i = 8$}
\put(-255,-8){\footnotesize$i = 9$}
\put(-123,-8){\footnotesize$i = 10$}
\put(-387,-78){\footnotesize$i = 11$}
\put(-255,-78){\footnotesize$i = 12$}
\put(-123,-78){\footnotesize$i = 13$}
\caption{\footnotesize Top: the estimated optimum over time. Blue and Red represent optimization with uniform and Gaussian drift proposals. Black lines indicate the local optima of the true functions. Bottom: a sequence of actions. Depicted are iterations 7-12 with uniform proposals.}\label{fig:bopt_results}
\end{figure}
In Fig. \ref{fig:bopt_results} we show results for our implementation of Bayesian Optimization 
with Thompson sampling. We compare two different proposal distributions, namely uniform proposals
and Gaussian drift proposals. We see that in this experiment, Gaussian drift is starting near 
the global optimum and drifts quickly towards it. (red curve, top panel of Fig. \ref{fig:bopt_results}).
Uniform proposals take longer to find the global optimum (blue curve, top panel of Fig. \ref{fig:bopt_results})
but we see that it can surpass the local optima of the curve\footnote{In fact, repeated experiments have
shown that when the Gaussian drift proposals starts near a local optimum, it gets stuck there. Uniform
proposals do not.}.
The bottom panel of Fig. \ref{fig:bopt_results} depicts a sequence of actions using uniform proposals.
The sequence illustrates the exploitation exploration trade-off that the implementation overcomes.
We start with complete uncertainty ($i=2$).
The Bayesian agent performs exploration until it gets a (wrong!) idea of where the optimum could be
(exploiting the local optima $i=5$ to $i=10$).
$i=11$ shows a change in tactic. The Bayesian agent, having exploited the local optima in previous
steps, is now reducing uncertainty in area it knows nothing about, eventually finding the global
optimum.

\section{Discussion}
This paper has shown that it is feasible and useful to embed Gaussian processes
in higher-order probabilistic programming languages by treating them as a kind
of statistical memoizer. It has described classic GP regression with both fully
Bayesian and MAP inference in a hierarchical hyperprior, as well as state-of-the-art
applications to discovering symbolic structure in time series and to Bayesian optimization.
All the applications share a common 50-line Python GP library and require fewer than 20 lines
of probabilistic code each.

These results suggest several research directions. First, it will be important
to develop versions of {\tt gpmem} that are optimized for larger-scale
applications. Possible approaches include the standard low-rank approximations
to the kernel matrix that are popular in machine learning~\citep{bui2014tree} as well
as more sophisticated sampling algorithms for approximate conditioning of the
GP~\citep{lawrence2009efficient}.
Second, it seems fruitful to abstract the notion of a ``generalizing" memoizer
from the specific choice of a Gaussian process model as the mechanism for
generalization. ``Generalizing" or statistical memoizers with custom regression techniques could be broadly useful in performance engineering and scheduling systems.
The timing data from performance benchmarks could be run through a generalizing memoizer by default.
This memoizer could be queried (and its output error bars examined) to inform the best strategy
for performing the computation or predict the likely runtime of long-running jobs.
 Third, the structure learning application suggests follow-on research in information
retrieval for structured data. It should be possible to build a time series search engine
that can handle search predicates such as ``has a rising trend starting around
1988" or ``is perodic during the 1990s".
The variation on the Automated Statistician presented in this paper can provide ranked result
sets for these sorts of queries because it tracks posterior uncertainty over structure and also
because the space of structural patterns that it can handle is easy to modify by making small
changes to a short VentureScript program.

The field of Bayesian nonparametrics offers a principled, fully Bayesian
response to the empirical modeling philosophy in machine learning~\citep{ghahramani2013bayesian},
where Bayesian inference is used to encode a state of broad ignorance rather
than a bias stemming from strong prior knowledge. It is perhaps surprising that
two key objects from Bayesian nonparametrics, Dirichlet processes and Gaussian
processes, fit naturally in probabilistic programming as variants of
memoization~\citep{roy2008stochastic}. It is not yet clear if the same will be true
for other processes, e.g. Wishart processes, or hierarchical Beta processes. We hope that the results in this paper encourage the development of other nonparametric libraries for higher-order probabilistic programming languages.

\myparagraph{Acknowledgements}
This research was supported by DARPA
  (under the XDATA and PPAML programs), IARPA (under research contract
  2015-15061000003), the Office of Naval Research (under research
  contract N000141310333), the Army Research Office (under agreement
  number W911NF-13-1-0212), the Bill \& Melinda Gates Foundation, and
  gifts from Analog Devices and Google.
\newpage
\section*{Appendix}
\subsection*{A Covariance Functions}
\begin{align}
\kse &= \sigma^2 \exp(-\frac{(x-x^\prime)^2}{2\ell^2}) \label{eq:SE}\\
\klin &=   \sigma^2(x x^\prime) \label{eq:LIN}\\
k^{\text{constant}} &=   \sigma^2\label{eq:C}\\
\kwn &= \sigma^2 \delta_{x,x^\prime} \label{eq:WN} \\
k^{\text{rational quadratic}}  &=    \sigma^2 \bigg(1 + \frac{(x - x^\prime)^2}{2 \alpha \ell^2} \bigg)^{-\alpha} \label{eq:RQ} \\
\kper &=  \sigma^2 \exp \bigg( \frac{2 \sin^2 ( \pi (x - x^\prime)/p}{\ell^2} \bigg). \label{eq:PER}
\end{align}
From top to bottom: the squared-exponential covariance function (\ref{eq:SE}),
also know as smoothing kernel; the linear kernel (\ref{eq:LIN});  the constant 
kernel (\ref{eq:C}); the white noise
kernel (\ref{eq:WN}); the
rational quadratic kernel (\ref{eq:RQ}); and the periodic kernel (\ref{eq:PER}).

\subsection*{B Covariance Simplification}
\begin{minipage}{\linewidth}
\small
\belowcaptionskip=-10pt
\begin{lstlisting}[frame=single,mathescape,label=alg:simplify,basicstyle=\selectfont\ttfamily]
SE $\times$ SE                  $\rightarrow$ SE 
{SE,PER,C,WN} $\times$ WN       $\rightarrow$ WN
LIN $+$ LIN                $\rightarrow$ LIN
{SE,PER,C,WN,LIN} $\times$ C    $\rightarrow$  {SE,PER,C,WN,LIN} 
\end{lstlisting}
\end{minipage}
Rule 1 is derived as follows:
\begin{equation}
\begin{aligned}
\sigma_c^2 \exp(-\frac{(x-x^\prime)^2}{2\ell_c^2})  &=  \sigma_a^2 \exp(-\frac{(x-x^\prime)^2}{2\ell_a^2}) \times  \sigma_b^2 \exp(-\frac{(x-x^\prime)^2}{2\ell_b^2}) \\
&= \sigma_c^2 \exp(-\frac{(x-x^\prime)^2}{2\ell_a^2}) \times   \exp(-\frac{(x-x^\prime)^2}{2\ell_b^2}) \\
&= \sigma_c^2 \exp \bigg(-\frac{(x-x^\prime)^2}{2\ell_a^2} -\frac{(x-x^\prime)^2}{2\ell_b^2}\bigg) \\
&= \sigma_c^2 \exp \bigg(-\frac{(x-x^\prime)^2}{2\ell_c^2}\bigg) \\
\end{aligned}
\end{equation}
For stationary kernels that only depend on the lag vector between $x$ and $x^\prime$ it holds that multiplying such a kernel with a WN kernel we get another WN kernel (Rule 2). Take for example the SE kernel:
\begin{equation}
 \sigma_a^2 \exp \bigg(-\frac{(x-x^\prime)^2}{2\ell_c^2}\bigg) \times  \sigma_b \delta_{x,x^\prime} =  \sigma_a \sigma_b \delta_{x,x^\prime}
\end{equation}
Rule 3 is derived as follows:
\begin{equation}
 \theta_c (x \times x^\prime) = \theta_a (x \times x^\prime) + \theta_b (x \times x^\prime) 
\end{equation}
Multiplying any kernel with a constant obviously changes only the scale parameter of a kernel (Rule 4).

\subsection*{C The Struct-Operator}

\begin{align*}
\Struct(\klin) &= \text{LIN}\\
\Struct(\kper) &= \text{PER}\\
\Struct(\kse) &= \text{SE}\\
\Struct(\kwn) &= \text{WN}\\
\Struct(k^{\text{linear+periodic}}) &= \text{LIN}+\text{PER}\\
\Struct(k^{\text{linear} \times \text{periodic}}) &= \text{LIN}
\times\text{PER}
\end{align*}

\newpage
\subsection*{D Glossary}
\begin{tabular}{l l}
$\mathcal{N}$		&  (Multivariate-) Gaussian \\
$\mathcal{GP}$		&  Gaussian Process \\
$\mathbb{E}$		&  Expectation    \\
$x,x_i$ 			&  Scalar, possibly indexed with $i$   \\
$\xbf$				&  Column vector, training data:
regression input (also actions in section \ref{sec:thompson})    \\
$\ybf$				&  Column vector, training data:
regression output  (also rewards in section \ref{sec:thompson})    \\
$\Acal$				& A set of possible actions \\

 $\xprime$			&  Column vector, unseen test input: regression input    \\
 $\yprime$			&  Column vector, sample from predictive posterior, that
is a sample from\\
&$\mathcal{N}(\mupost,\Kpost)$    \\

$\xstar$         &  Column vector, unseen test
input: regression input before any data\\
&has been observed    \\
$\ystar$		&  Column vector, sample from the predictive prior
conditioned on $\thetabf$\\
& and unseen test input $\xstar$ \\
$\Dbf$		&  Data matrix $[\xbf\; \ybf]$ \\

$\mu(x)$                 &  Mean function \\
$\thetabf_{\text{mean}}$	&  hyper-parameters for a mean function
\\
$k \text{ or } k(x_i,x_j)$	&  a covariance function or kernel, that is a
function that takes two scalars as input \\

$\thetabf$			&  hyper-parameters for a
kernel/covariacne function (also semicontext in section \ref{sec:thompson})\\
$k(x_i,x_j \midtheta )$		&  a kernel conditioned on its
hyper-parameters \\
$K(\xbf,\xprime \midtheta )$    &  Function outputting a matrix of dimension $I \times J$
with entries $k(x_i,x_j \midtheta )$;\\
&with $x_i \in \xbf$ and $x_j \in \xprime$ where $I$ and $J$ indicate the length
of the\\
&column vectors $\xbf$ and
$\xprime$ \\

$\ktheta$			&  a covariance function parameterized with
$\thetabf$\\

$\Ktheta\text{ or }\Kbf_{(\thetabf,\xbf,\xbf)}$		&  covariance matrix
computed with by $K(\xbf,\xbf \midtheta)$   \\
$\mupost$ &  Posterior mean vector for $\yprime \mid \xbf, \xprime, \ybf, \thetabf$ \\
$\Kpost$     &  Posterior covariance matrix for $\yprime \mid \xbf, \xprime, \ybf, \thetabf$ \\
$\Lbf$			&lower triangular matrix, given by the Cholesky factorization as 
$\Lbf\coloneqq \text{chol}(\Ktheta)$\\

$\kse$ &  Squared exponential covariance function  \\
$\klin$ &  Linear covariance function  \\
$k^{\text{constant}}$ &  Constant covariance function  \\
$\kwn$ &  White noise covariance function  \\
$k^{\text{rational quadratic}}$  &  Rational quadratic covariance function  \\
$\kper$ &  Periodic covariance function  \\

SE			&  Symbolic expression for the squared exponential covariance function  \\
LIN		        &  Symbolic expression for the linear covariance function  \\
PER		        &  Symbolic expression for the periodic covariance function  \\
RQ		        &  Symbolic expression for the rational quadratic covariance function  \\
C		        &  Symbolic expression for the constant covariance function  \\
WN		        &  Symbolic expression for the white noise covariance function  \\
\end{tabular}
\newpage
\begin{tabular}{l l}
$\land$                &  Logical and \\
$\lor$                 &  Logical or \\
$\Krv$			& Random variable over kernel functions \\
kernel functions \\
Parse$(k)$		& Parse the structure for kernel $k$ \\
Simplify$(k)$		& Simplify the functional expression for kernel $k$ \\
Struct$(k)$		& Symbolic interpretation for kernel $k$ \\
$\Cont(k,k^t)$          & A kernel $k^t$ contains the global kernel structure
$k$\\
$\SEK$			&  Operator to check if $\Struct(\Ksrv=k)=\text{SE}$\\
$\LINK$			&  Operator to check if $\Struct(\Ksrv=k)=\text{LIN}$\\
$\PERK$			&  Operator to check if $\Struct(\Ksrv=k)=\text{PER}$\\
$\WNK$			&  Operator to check if $\Struct(\Ksrv=k)=\text{WN}$\\
BK			& A set of base kernels \\
$\Sbf$			& A subset of BK, randomly selected \\
$\Omegabf$		& Random variable for composition operators, in our
case that is kernel addition\\
&and multiplication $\{+,\times\}$\\
$\Gamma(\alpha,\beta)$ & Gamma distribution with shape parameter $\alpha$ and
rate $\beta$ \\

$\ell$			& Length-scale parameter for $\kse$  \\
$sf$			& Scale factor parameter\\
$\ttheta \in \Theta$    & Context in Thompson sampling\\ 
$\Theta$    & Context space\\ 
$V$			& Value function\\
$Q_\proposal$		& Proposal distribution\\
$E$			& Energy function\\
$s$			& Temperature parameter \\
\end{tabular}


\newpage
\bibliography{VentureGP.01}
\bibliographystyle{apalike}
\end{document}